\documentclass{article} 
\usepackage[preprint]{colm2025_conference}

\usepackage{microtype}
\usepackage{hyperref}
\usepackage{url}
\usepackage{booktabs}
\usepackage{wrapfig}
\usepackage{times}
\usepackage{latexsym}
\usepackage{listings}
\usepackage{adjustbox}
\usepackage{amsmath}
\usepackage{amssymb}
\usepackage{amsthm}
\usepackage{graphicx}
\usepackage{subcaption} 
\usepackage{float}
\usepackage{afterpage}
\usepackage[skip=2pt]{caption}
\usepackage{multirow}
\usepackage{booktabs}
\usepackage{float} 
\usepackage{caption}
\usepackage{tabularx} 

\usepackage{CJKutf8}
\newcommand{\chinese}[1]{\begin{CJK}{UTF8}{gbsn}#1\end{CJK}}

\theoremstyle{assumption}
\newtheorem{assumption}{Assumption}

\theoremstyle{theorem}
\newtheorem{theorem}{Theorem}

\theoremstyle{corollary}
\newtheorem{corollary}{Corollary}

\usepackage{lineno}

\definecolor{darkblue}{rgb}{0, 0, 0.5}
\hypersetup{colorlinks=true, citecolor=darkblue, linkcolor=darkblue, urlcolor=darkblue}

\title{How Well do LLMs Compress Their Own Chain-of-Thought? \\ A Token Complexity Approach}

\author{Ayeong Lee\thanks{Equal contribution.} \\ Columbia Business School \\ \texttt{al3393@columbia.edu} \\\And Ethan Che$^{*}$ \\ Columbia Business School \\ \texttt{ewc2119@columbia.edu} \\\And Tianyi Peng \\ Columbia Business School \\ \texttt{tp2845@columbia.edu}}

\begin{document}

\ifcolmsubmission
\linenumbers
\fi

\maketitle

\begin{abstract}

Chain-of-thought prompting has emerged as a powerful technique for enabling large language models (LLMs) to solve complex reasoning tasks. However, these reasoning chains can be verbose, raising concerns about efficiency. In response, recent works have sought to decrease response lengths through simple prompting strategies (e.g. `be concise'). In this work, we conduct the first systematic study of the relationship between reasoning length and model performance across a diverse range of compression instructions (e.g. `use 10 words or less' or 'remove all punctuation'). In doing so, we discover a universal tradeoff between reasoning length and accuracy that persists across even very distinct reasoning chains. We demonstrate that this tradeoff emerges from a sharp threshold behavior at the question level: each task has an intrinsic `token complexity' – a minimal number of tokens required for successful problem-solving. We show how token complexity enables us to compute upper bounds on the optimal accuracy-compression tradeoff. Our analysis reveals that prompt-based compression strategies operate far from these theoretical limits, suggesting significant room for improvement and providing benchmarks to help researchers evaluate progress in reasoning efficiency. Our work also highlights the importance of \emph{adaptive compression} -- giving shorter responses for easier questions -- and we show that token complexity is a useful tool for measuring this capability.
\end{abstract}

\section{Introduction}

Recent advancements in large language models (LLMs)—including o1 and DeepSeek R1—alongside broader AI agent development, have showcased impressive reasoning capabilities, hinting at a future where complex problem-solving and decision-making can be automated. 
However, this rapid progress also introduces a significant challenge: the computational cost of reasoning is projected to increase substantially as these models are deployed in real-world applications. This growing inference cost highlights the necessity for efficient reasoning strategies, motivating our research into reducing the computational burden of LLM inference while maintaining high performance.

A pivotal technique for enhancing LLM reasoning has been chain-of-thought (CoT) prompting, which encourages models to generate intermediate reasoning steps before arriving at a final answer~\citep{wei2022chain, kojima2022large}. While effective, these reasoning chains often involve lengthy intermediate computations, increasing inference costs when deployed at scale~\citep{yu2024distilling}.

Although CoT-generated outputs are typically verbose, containing redundant~\citep{chiang2024over} or sometimes irrelevant information~\citep{wang2023towards}, it remains unclear how best to compress the chain-of-thought content for effective problem-solving.  Prior work has observed that asking the LLM to `be concise' or `use at most 100 words' can reduce response length while incurring a range of degradation to accuracy~\citep{jin2024impact, renze2024benefits, han2024token}. Yet so far, there has not been a comprehensive exploration of the effectiveness of different compression strategies. For example, should language models use fewer reasoning steps instead of using fewer words or characters? Should they be removing unnecessary punctuation or filler words (e.g. `therefore', `so'). Should they reason in alternative languages that may be more token-efficient?

\begin{table*}[t]
  \centering
  \footnotesize 
  \renewcommand{\arraystretch}{1.} 
  \begin{tabularx}{\textwidth}{l p{4.8cm} >{\raggedright\arraybackslash}X} 
    Prompt Id           & Prompt & Sample Output under Claude 3.5 Sonnet\\
    \hline
       \textsf{NoCoT}            & Only give the final answer. & \textit{Answer: (G)} \\
    	\textsf{DefaultCoT}       & Think step-by-step. & \textit{Let me solve this step by step: 1) Z18 (...)} \\
    	\textsf{BeConcise}        & Be concise. & \textit{Let's solve this step by step: 1) Z18 (...)} \\
    	\textsf{BulletPoints}     & Only use bullet points. & \textit{- Z18 is the group of integers modulo (...)} \\
   	 \textsf{OnlyNumbers}      & Only use numbers or equations. & \textit{|Z18| = 18. Answer: (G)} \\
    	\textsf{NoSpaces}         & Do not use any spaces or line breaks. & \textit{TheorderofagroupZnisthesame (...)}\\
   	 \textsf{NoProperGrammar}  & Do not use proper grammar. & \textit{lemme help u with this z18 is group} \\
   	 \textsf{AbbreviateWords}  & Abbreviate words as much as possible. & \textit{Slvng fr rdr f Z18: Z18 = grp f ntgrs (...)} \\
   	 \textsf{WordLimit(k)}     & Use at most $k$ words. $(k \in [1,100])$ & \textit{Order of Z18 is eighteen. Answer: (G)} \\
   	 \textsf{CharLimit(k)}     & Use at most $k$ letters.  $(k \in [1,500])$& \textit{The order of a group is the number of (...)} \\
    	\textsf{TokenLimit(k)}    & Use at most $k$ tokens.   $(k \in [1,500])$ & \textit{The order of Z18 is 18, as it contains (...)} \\
   	 \textsf{StepLimit(k)}     & Use at most $k$ steps. $(k \in [1,5])$  & \textit{Step 1: The order of a group Zn is  (...)} \\
      \textsf{ChineseCoT}    & Respond in Chinese  & \chinese{让我帮你解答这个问题} (...)\\
  	  \textsf{ChineseCoT(k)}    & Use at most $k$ Chinese characters.   & \chinese{答案是18}. Answer: (G).
       \\
    \hline
  \end{tabularx}
  \caption{\label{tab:prompts} The Chain-of-Thought prompts we consider. The right column gives an example of the chain-of-thought of Claude 3.5 Sonnet on a sample problem in MMLU-Pro Math. We indicate the range of $k$ that we consider.}
  \label{tab:commands}
\end{table*}

The main contribution of our work is to provide the first systematic study of the trade-off between reasoning length and performance across different prompt-based compression strategies, including prior strategies such as `be concise' as well as alternative approaches such as `only use bullet points' or `use at most 50 Chinese characters'. In total, we evaluate 31 prompts for six LLMs on three standard reasoning datasets. Remarkably, although these prompting strategies produce  different chains of thought, their trade-offs between response length and accuracy lie on a universal trade-off curve. In other words, all prompts are equally "good" as extremes on this curve. What primarily affects accuracy is the \emph{length} of the chain of thought, far more than changes in its composition.


Our second contribution is a novel empirical observation: the performance of reasoning tasks exhibits a sharp threshold dependence on reasoning length at the \emph{question} level. By evaluating multiple prompts for each question, we demonstrate that most questions have a well-defined `token complexity'—a minimum number of tokens required to successfully solve the question--which holds across diverse prompting strategies. We estimate token complexities across various benchmarks and find that:
(i) Token complexity alone can predict the performance of CoT prompting strategies with 94\% accuracy.
(ii) It serves as a robust measure of reasoning task difficulty, enabling us to investigate whether LLMs reason \emph{adaptively}—using shorter chains-of-thought for easier questions.

These results raise the question of whether the trade-off curve between response-length and accuracy induced by these prompting strategies are close or far from optimal. Viewing these strategies as a form of `lossy compression', we take inspiration from rate-distortion theory to characterize an upper bound on the optimal accuracy-compression trade-off. In doing so, we find that prompt-based strategies are far from this upper bound, especially on harder datasets. 
We also illustrate how our experimental results can be used to evaluate methodologies for improving the adaptivity of LLM reasoning. We show evidence that even without explicitly prompting, LLMs do exhibit adaptive reasoning lengths and that the simple prompting strategies we test offer a surprisingly strong baseline that can outperform more sophisticated prompt-routing strategies. Finally, we show that if one has access to a verifier, one can achieve a substantially better accuracy-length tradeoff and approach the upper bound.

\subsection{Related Work}
Recent research has begun to gain traction in exploring the trade-off between response length and accuracy in LLMs. Studies such as~\citep{renze2024benefits, jin2024impact, nayab2024concise, han2024token} have employed specific prompting strategies to constrain response length and assess the associated impact on accuracy and performance. Additionally, several works have highlighted the redundancy inherent in CoT prompting~\citep{chiang2024over, wu2025more} and emphasized the benefits of concise reasoning. Other approaches have focused on fine-tuning strategies to adapt LLMs for generating more succinct reasoning~\citep{kang2024c3ot, yu2024distilling}.

Our work advances this growing body of literature by making three key contributions: (1) we conducted a systematic evaluation of a rich set of prompts designed to reduce the length of CoT reasoning while maintaining accuracy. (2) We find that accurate answers are only achieved when the output length exceeds a certain threshold, which is intrinsic to the problem and independent of the CoT format. We formalize this concept as the \textit{token complexity} of a problem. (3) We derive theoretical limits on the length-accuracy tradeoff, providing a framework for researchers to benchmark new methodologies aimed at compressing chain-of-thought reasoning effectively.

\begin{figure}[t]
  \centering

    \begin{minipage}{\textwidth}
  \includegraphics[width=0.50\textwidth]{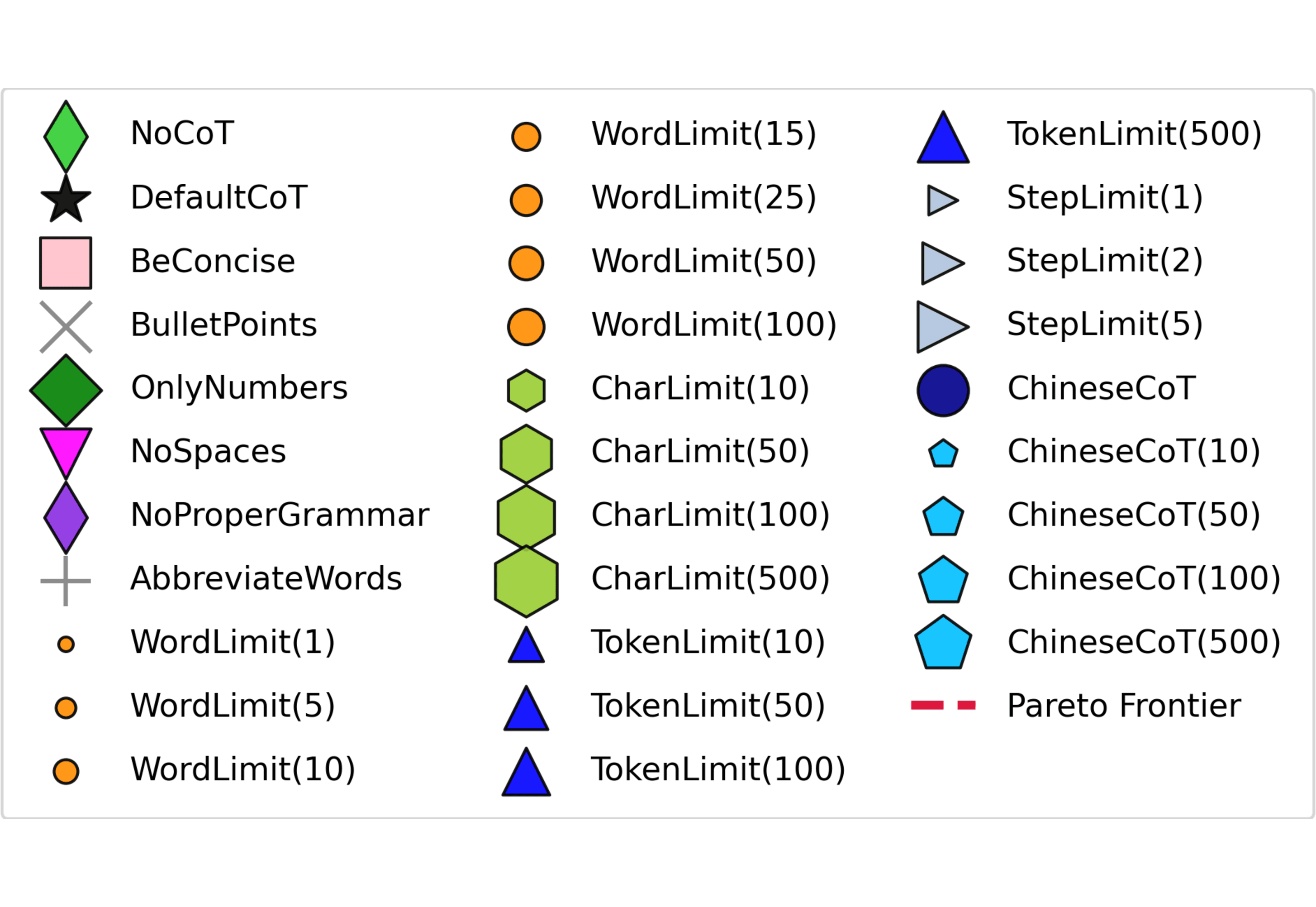}
  \includegraphics[width=0.5\textwidth]{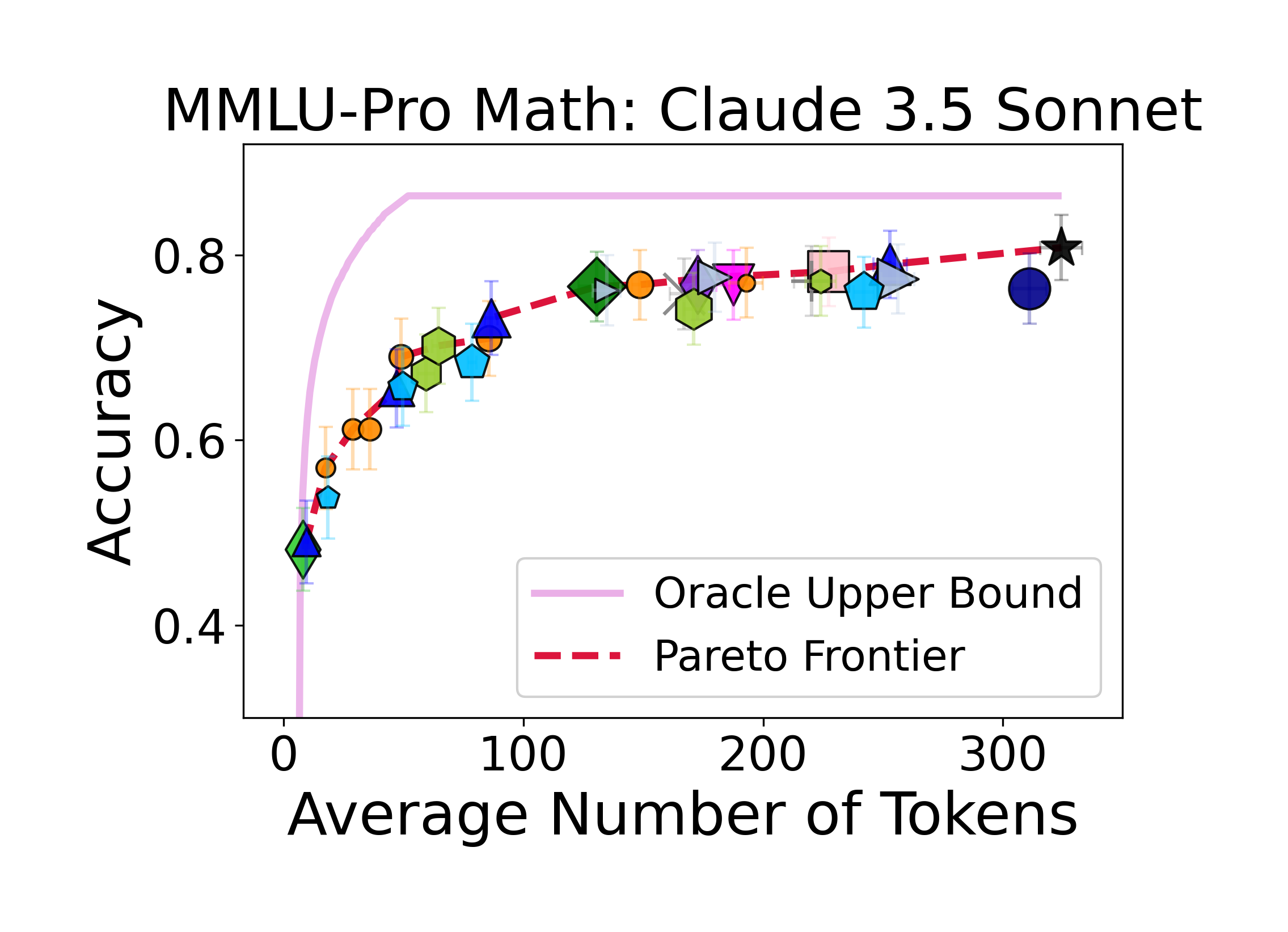}
  \end{minipage}

  \begin{minipage}{\textwidth}
  \includegraphics[width=0.5\textwidth]{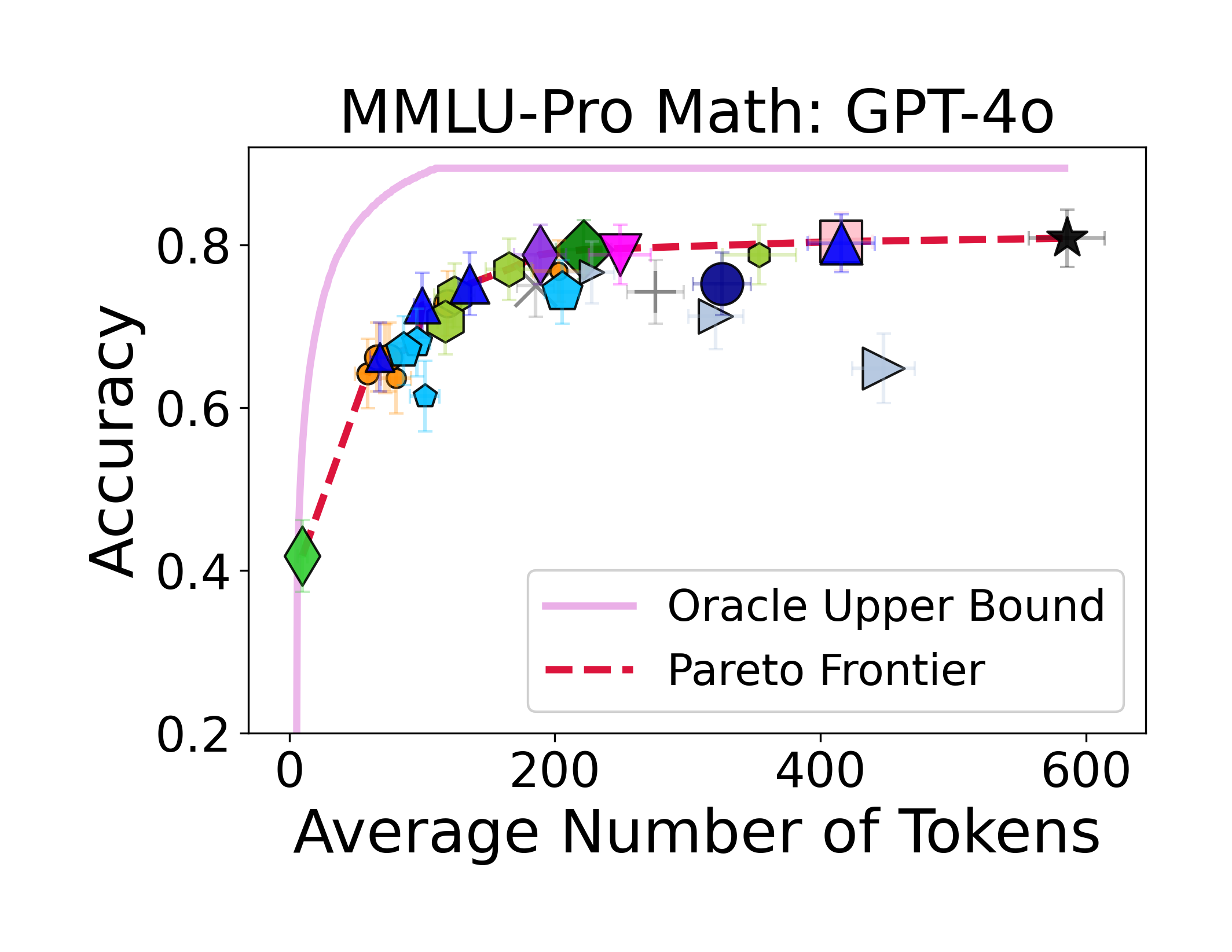}  
  \includegraphics[width=0.5\textwidth]{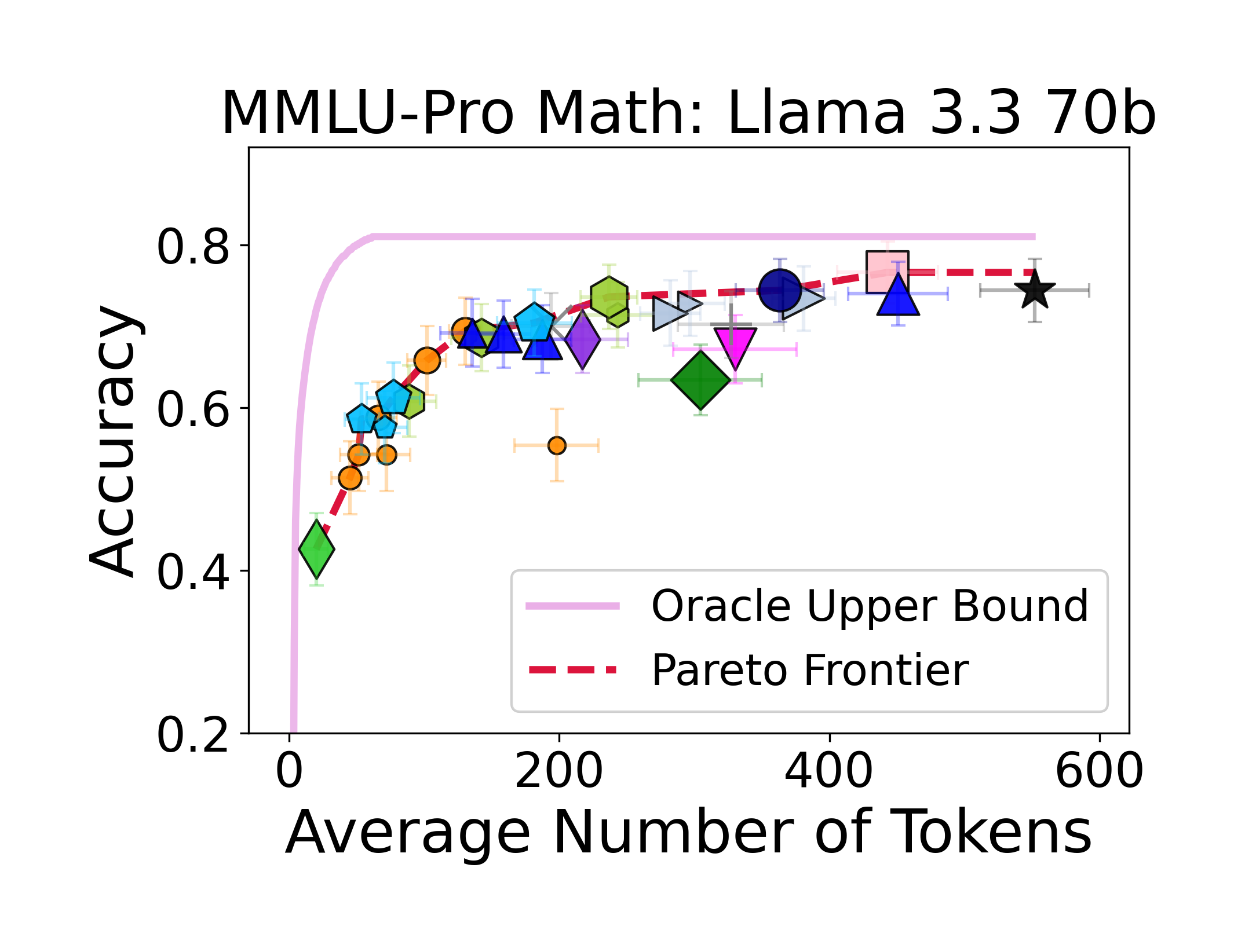}
  \end{minipage}
  
  \caption{\label{fig:tradeoff} For each of the 31 CoT prompts we consider (see legend above), we report average token length vs accuracy for GPT-4o and Claude 3.5 Sonnet on MMLU-Pro Math and GPT-4o-mini on GSM8K tasks. Despite differences in the chain-of-thought, many of them live on a universal tradeoff curve. Using our framework, we compute upper bounds on optimal accuracy under a given average token budget (see section~\ref{sec:bounds}). See Appendix~\ref{sec:appendix_tradeoff} for more models and benchmarks.}
\end{figure}

\section{Experiments}
Our evaluation encompasses the following LLMs: GPT-4o \citep{hurst2024gpt}, GPT-4o-mini~\citep{hurst2024gpt}, Claude 3.5 Sonnet~\citep{anthropic2024claude}, Claude 3.5 Haiku~\citep{anthropic2024claude}, and Llama 3.3 70B Instruct~\citep{dubey2024llama}. We evaluate these models on three standard math reasoning datasets: MATH-500~\citep{lightman2023lets}, a random 500 problem subset of GSM8K~(GSM8K, \cite{cobbe2021training}), and a random 500 problem subset of MMLU-Pro Math problems~(MMLU-Pro Math, \cite{wang2024mmlu}).

For each LLM and dataset, we test 31 prompts designed to induce shorter response lengths, detailed in Table~\ref{tab:prompts}. These prompts include ones considered in prior literature: `be concise'~\citep{renze2024benefits}, `use $k$ words or less'~\citep{jin2024impact, nayab2024concise}, `use $k$ tokens or less'~\citep{han2024token}, but include additional curated ones to assess the impact of alternative compression strategies. For each prompt, we assess performance with two metrics: (1) {\bf accuracy}, the fraction of questions solved correctly, and (2) {\bf average token length}, the average number of output tokens produced by the LLM in their response across questions in the dataset.

\subsection{Benchmark Results}

By considering multiple diverse prompts, we are able to induce chains-of-thought along a spectrum of response lengths and reasoning performance, with \textsf{NoCoT} (no chain-of-thought) using the fewest tokens with the lowest accuracy and \textsf{DefaultCoT} (i.e. standard chain-of-though-prompting `think step-by-step') using the most tokens and generally having the highest benchmark performance. 
In Table~\ref{tab:mmlu}, we focus on a subset of chain-of-thought prompts and report the accuracy and average token count for MMLU-Pro Math across several LLMs. We highlight the following observations, which also hold for other datasets (see Appendix~\ref{sec:appendix_tables}).

\begin{itemize}
    \item There is potential to achieve significant length reduction (e.g., up to 60\%) compared to \textsf{DefaultCoT} without sacrificing much accuracy.
    \item \textsf{BeConcise}~\citep{renze2024benefits} indeed consistently reduces token length without significantly hurting performance. 
    \item Yet, there are other prompts such as \textsf{OnlyNumbers} or \textsf{NoProperGrammar} that preserve a similar accuracy as \textsf{BeConcise} but induce shorter chains-of-thought (a $\approx50\%$ reduction for GPT-4o). The strong performance of these prompts suggests that LLMs do not necessarily require proper English to conduct chain-of-thought reasoning effectively.
    \item There is no universally dominant prompt; while \textsf{OnlyNumbers} and \textsf{NoProperGrammar} are effective for GPT-4o and Claude 3.5 Sonnet, they are less so for LLaMA 3.3 70B.
    \item While LLMs do not always follow the prompt exactly, the prompt still induces large variations in the response length (as also observed in ~\cite{han2024token}).
\end{itemize}

As researchers (e.g. \cite{kang2024c3ot}) propose new methodologies for improving the accuracy-length trade-off, our experimental results provide a simple yet effective baseline.

\subsection{Universal Trade-off between Reasoning Length and Accuracy}
\label{sec:tradeoff}

The observation that the token length of \textsf{DefaultCoT} can be substantially improved upon without much degradation to accuracy motivates a natural question: which prompts exhibit the best tradeoff between response length and accuracy?

\begin{wraptable}{r}{0.6\textwidth}
\vspace{-1em}
\centering
\scriptsize
\begin{tabular}{lcccccc}
\toprule
Prompt 
& \multicolumn{2}{c}{GPT-4o} 
& \multicolumn{2}{c}{Claude 3.5 Sonnet} 
& \multicolumn{2}{c}{Llama-3.3-70B} \\
& Acc & Tok & Acc & Tok & Acc & Tok \\
\midrule
\textsf{NoCoT}           & 42\% & 10  & 48\% & 8   & 43\% & 20  \\
\textsf{DefaultCoT}      & 81\% & 586 & 81\% & 324 & 74\% & 552 \\
\textsf{BeConcise}       & 80\% & 415 & 78\% & 227 & 77\% & 443 \\
\textsf{BulletPoints}    & 75\% & 185 & 76\% & 167 & 70\% & 194 \\
\textsf{OnlyNumbers}     & 79\% & 222 & 77\% & 130 & 63\% & 304 \\
\textsf{NoProperGrammar} & 79\% & 189 & 77\% & 173 & 68\% & 217 \\
\textsf{ChineseCoT}      & 75\% & 326 & 76\% & 311 & 74\% & 363 \\
\bottomrule
\end{tabular}
\caption{Comparison of Accuracy and Average Token Length of chain-of-thought prompts on MMLU-Pro Math. }
\label{tab:mmlu}
\vspace{-1em}
\end{wraptable}
To study this question, we plot the average token-length and accuracy of all 31 prompts in Figure~\ref{fig:tradeoff} for the MMLU-Pro Math benchmark (results for other benchmarks are in Appendix~\ref{sec:appendix_tradeoff}).
Remarkably, we see that almost all the prompts we consider lie on a \emph{universal} trade-off curve between response length and accuracy. This suggests that regardless of whether the chain-of-thought is formatted in bullet points, without spaces, using only numbers, or even in Chinese, ultimately it is the length of the chain-of-thought that matters most.
This result also holds for the \textsf{Wordlimit(k)}, \textsf{Charlimit(k)}, \textsf{TokenLimit(k)}, \textsf{StepLimit(k)}, and \textsf{ChineseCoT(k)} prompts, which ask the LLM to limit the response to be at most $k$ words, letters, tokens, reasoning steps, or Chinese characters. One would expect that each of these strategies would result in separate tradeoff curves, but surprisingly all of these strategies result in a near-identical trade-off between response length and accuracy. This suggests that there is not much room for improvement by changing the composition of the chains-of-thought alone.

This universal trade-off curve suggests that the length of the chain-of-thought is the predominant factor that influences reasoning performance. We caveat this observation acknowledging that this universal trade-off should only hold for reasonably informative chains-of-thought, i.e. we would expect that pure white-space would perform worse. We also see that adherence to the universal trade-off curve is better for more capable models (i.e. GPT-4o and Claude-3.5-Sonnet) on easier benchmarks (i.e. GSM8K). For less capable models such as LLaMA 3.3 70B on harder datasets (e.g. MATH-500), there are more prompts which are below the trade-off curve.

\section{The Token Complexity Hypothesis}
\label{sec:token-complexity}

The results from the previous section highlight the importance of response length on reasoning performance. To investigate this relationship further, we use our dataset to study reasoning performance at question-level granularity. Our extensive coverage of response lengths allows us to observe that reasoning performance at the question-level exhibits a sharp threshold-like behavior: across all prompts, the LLM correctly solves the question if and only if the response length is above a certain threshold. We refer to this threshold as the \emph{token complexity} of the problem.
  \begin{figure}[t]
    \centering
    \hspace{-1.5cm} 
    \begin{minipage}{0.55\textwidth} 
      \centering
      \includegraphics[width=\linewidth]{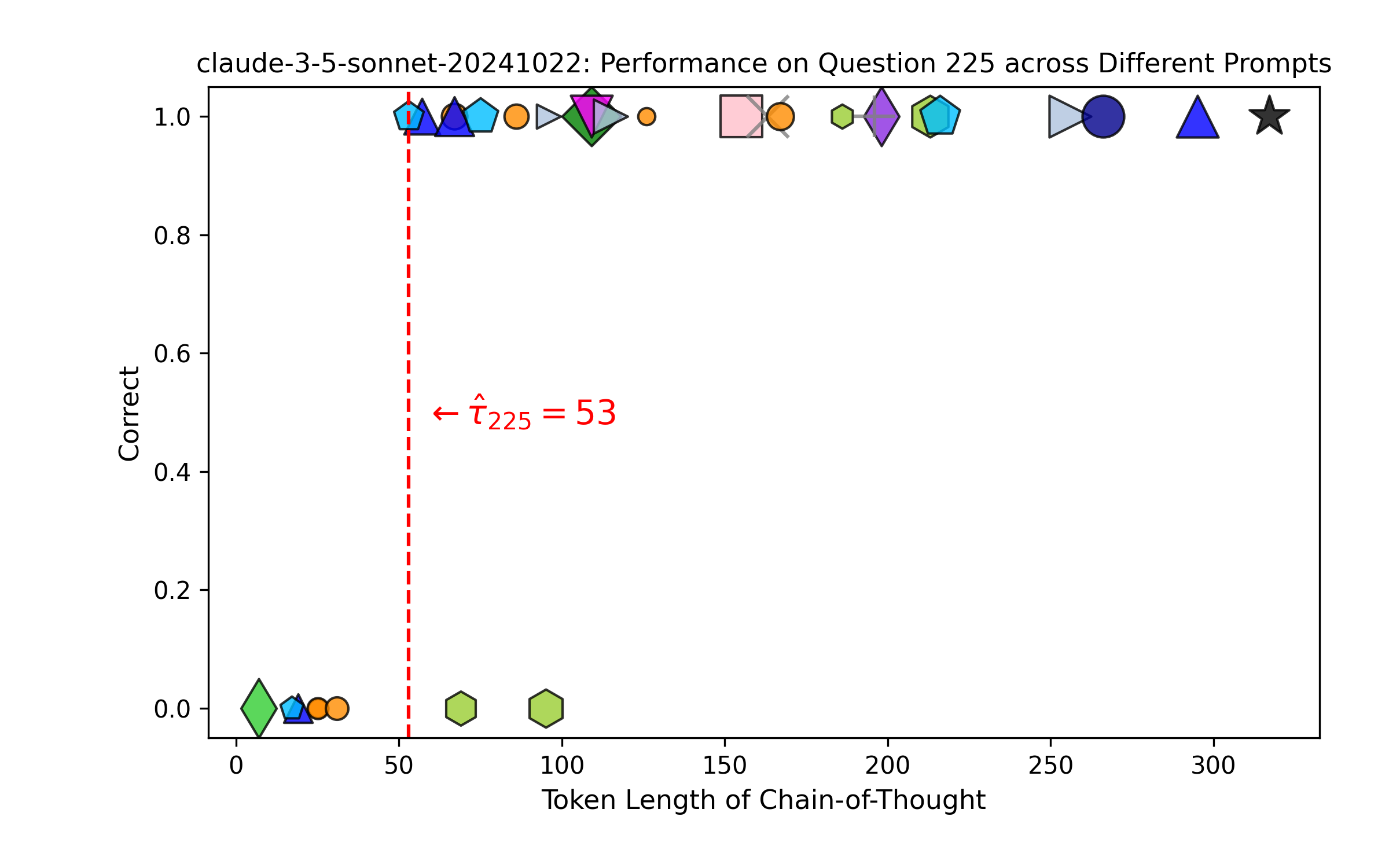}
    \end{minipage}
    \hfill
    \begin{minipage}{0.48\textwidth} 
      \centering
            \includegraphics[width=\linewidth]{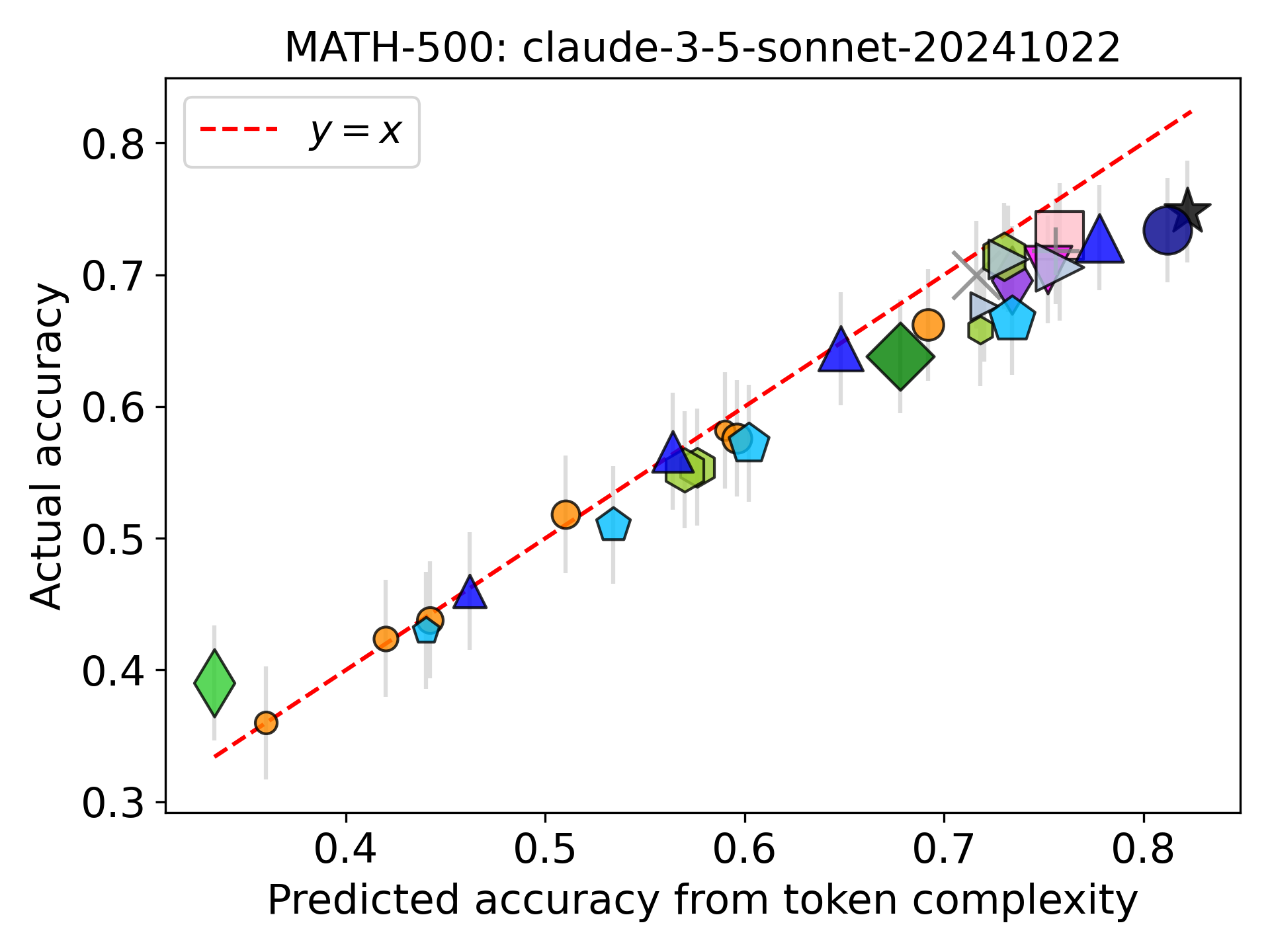}
    \end{minipage}
          \caption{\label{fig:token-complexity} Illustration of the token complexity hypothesis. (Left) Performance of Claude 3.5 Sonnet on a sample question in MATH-500 exhibits a threshold behavior. With the exception of 2 prompts, all the prompts that use more tokens than the threshold get the answer correct, the rest do not. The red dotted line indicates the estimated token complexity $\tau_{i}^{\pi}$ from the results. (Right) Actual benchmark accuracy on MATH-500 vs predicted accuracy from the token complexity hypothesis. Token complexity is highly predictive of actual accuracy (see Table~\ref{tab:err}).}
  \end{figure}

To illustrate, the left panel of Figure~\ref{fig:token-complexity} displays the performance of all 31 prompts for Claude 3.5 Sonnet on a sample question in the MATH-500 dataset. Despite the diversity of prompting strategies we consider, we see that response length is highly predictive of correctness: with the exception of 2 prompts, all the prompts which use more than $\approx 53$ output tokens correctly solve the problem, while the prompts that use fewer tokens get the question wrong.

To formalize this behavior, 
we first introduce some notation. Given a dataset of $i=1,...,n$ questions ($n = 500$ for the benchmarks we consider), let $P_{k}$ denote a chain-of-thought prompt and let $X_{i,k}$ denote a chain-of-thought produced by an LLM $\pi$ for question $i$ when prompted by $P_{k}$. We let $t(X_{i,k}) \in \mathbb{N}$ be the length of $X_{i,k}$ in tokens.  We let $a^{\pi}_{i}(X_{i,k}) = 1$ if $\pi$ gets the answer correct under $X_{i}$ and $0$ if not. We now turn to formally outlining the `token complexity hypothesis'.


\begin{assumption}
\label{ass:token-complexity}
(Token complexity hypothesis) For question $i$ and LLM $\pi$, there exists a threshold $\tau^{\pi}_{i}\in \mathbb{N}$, denoted as the \textbf{token complexity}, such that for any prompt $P_{k}$, the LLM gets the answer correct iff the token length $t(X_{i,k})$ is above $\tau^{\pi}_{i}$:
\[
a^{\pi}_{i}(X_{i,k}) = \mathbf{1}\{t(X_{i,k}) \geq \tau^{\pi}_{i} \}
\]
\end{assumption}
As discussed before, we should think of this hypothesis as holding only for `reasonable' chains-of-thought (e.g. not 100 tokens of pure whitespace); and we observe this to hold broadly for all the prompts we test.
This universality makes the token complexity notion useful because it provides fine-grained notion of problem difficulty that doesn't depend on a specific prompting strategy: more difficult problems require more tokens to solve.

\begin{table}[t]
  \centering
  \begin{tabular}{lccccccccc}
\toprule
Model
& \multicolumn{2}{c}{MMLU-Pro Math} 
& \multicolumn{2}{c}{GSM8K} 
& \multicolumn{2}{c}{MATH-500} \\
& $\bar{c}_{\pi}$ & $\text{Err}_{\pi}$ 
& $\bar{c}_{\pi}$ & $\text{Err}_{\pi}$ 
& $\bar{c}_{\pi}$ & $\text{Err}_{\pi}$ \\
\midrule
GPT-4o              & 92.1\% & 5.0\% & 97.0\% & 1.6\% & 89.9\% & 6.0\% \\
GPT-4o Mini         & 91.3\% & 5.2\% & 94.1\% & 5.1\% & 91.4\% & 5.7\% \\
Claude 3.5 Sonnet   & 92.6\% & 3.4\% & 97.4\% & 1.4\% & 92.1\% & 4.5\% \\
Claude 3.5 Haiku    & 90.1\% & 5.1\% & 94.8\% & 4.1\% & 90.2\% & 6.3\% \\
Llama 3.3 70B       & 90.6\% & 5.1\% & 96.2\% & 2.7\% & 89.8\% & 6.1\% \\
\bottomrule
\end{tabular}

  \caption{\label{tab:err} Evidence of token complexity hypothesis across LLMs and reasoning benchmarks. 
  $\bar{c}_{\pi}$ is the average classification accuracy of the threshold classifier $\mathbf{1}\{t(X_{i,k}) > \hat{\tau}_{i}^{\pi}\}$ in predicting whether the LLM gets the question correct or not.
   $\text{Err}_{\pi}$ is the discrepancy between actual performance $\text{Acc}_{\pi}(P_k)$ and 
  predicted performance $\widehat{\text{Acc}}_{\pi}(P_k)$ from the token complexity hypothesis. The high classification accuracy of $\bar{c}_{\pi} \approx 90\%$ and the small discrepancy $\text{Err}_{\pi} \approx 6\%$ illustrate that the token complexity hypothesis has strong predictive power for the reasoning performance of LLMs at both the question and benchmark levels.}


\end{table}

We proceed to test this hypothesis quantitatively across LLMs $\pi$ and benchmarks. First, we measure to what degree success or failure on a task can be classified based purely on the token-length of the chain-of-thought, i.e. whether the behavior in the left panel of Figure~\ref{fig:token-complexity} holds broadly. To do so, we use our dataset of $k=1,...,K=31$ chain-of-thought prompts for each question. For a threshold $t\in \mathbb{N}$, we define $c^{\pi}_{i}(t)$ to be the classification accuracy under a threshold classifier under $t$, which measures how predictable reasoning success is based on whether the token count exceeds threshold $t$
\[
c^{\pi}_{i}(t) \equiv \frac{1}{K}\sum_{k=1}^{K}\mathbf{1} \{ a_{i}^{\pi}(X_{i,k}) = \mathbf{1}\{t(X_{i,k}) \geq t \}\}
\]
Our estimator $\hat{\tau}_{i}^{\pi}$ of token complexity is the optimal threshold-based classifier, and we let $c_{i}^{*}$ be the maximum classification accuracy achieved by $\hat{\tau}^{\pi}_{i}$. 
\begin{align}
\hat{\tau}^{\pi}_{i} &\equiv \arg\max_{k} c_{i}^{\pi}(t(X_{i,k})), \quad c_{i}^{*} \equiv \max_{k} c_{i}^{\pi}(t(X_{i,k})),
\quad\bar{c}_{\pi}\equiv \frac{1}{n}\sum\nolimits_{i=1}^{n} c_{i}^{*}
\end{align}
We let $\hat{\tau}_{i}^{\pi} = \infty$ if setting the threshold to $t=\infty$ results in better classification accuracy, e.g. if none of the chains-of-thoughts correctly solve the problem in which case $a_{i}^{\pi}(X_{i,k}) = 0$ for all $k$. For each dataset and model, we report the average classification accuracy $\bar{c}_{\pi}$ under the estimated $\tau^{\pi}_{i}$ in Table~\ref{tab:token-limit}. Overall, the average classification accuracy is very high, above 90\% for most models and benchmarks, verifying that (1) token-length is highly predictive for performance at the question-level and (2) question-level performance exhibits a threshold relationship with token-length. We also see that the threshold classifier achieves higher average classification accuracy for (1) more capable models (i.e. GPT-4o and Claude 3.5 Sonnet) and (2) easier benchmarks (i.e. GSM8K), which mirrors the finding in Section~\ref{sec:tradeoff} concerning adherence to the universal trade-off curve.

Our second empirical validation is to compare the accuracy of the LLM across prompts and benchmarks with the performance predicted under the token complexity hypothesis. We let $\text{Acc}_{\pi}(P_{k})$ denote the accuracy of LLM $\pi$ under prompt $P_{k}$ while $\widehat{\text{Acc}}_{\pi}(P_{k})$ denotes the accuracy predicted under our estimated token-complexities:
\begin{align}
\text{Acc}_{\pi}(P_{k}) \equiv \frac{1}{n}\sum\nolimits_{i=1}^{n} a^{\pi}_{i}(X_{i,k}),\quad 
\widehat{\text{Acc}}_{\pi}(P_{k}) &\equiv \frac{1}{n}\sum\nolimits_{i=1}^{n} \mathbf{1}\{t(X_{i,k}) \geq \hat{\tau}^{\pi}_{i} \}
\end{align}
In Table~\ref{tab:err}, for each LLM and dataset we report $\text{Err}_{\pi}$, the average relative discrepancy between actual accuracy and predicted accuracy:
\[
\text{Err}_{\pi} = \frac{1}{K}\sum\nolimits_{k=1}^{K} \frac{| \text{Acc}_{\pi}(P_{k}) - \widehat{\text{Acc}}_{\pi}(P_{k}) |}{\text{Acc}_{\pi}(P_{k})}
\]
We observe that the token threshold classifier is able to predict overall benchmark performance within 6\% error, and the error is even smaller for larger models on easier benchmarks. This illustrates that the token-complexity hypothesis provides an accurate model for reasoning performance, which can be seen visually in the right panel of Figure~\ref{fig:token-complexity} for Claude 3.5 Sonnet on the MATH-500 dataset and helps characterize the strong dependence of token-length on accuracy.
We explore the implications of this hypothesis in the next section.

\begin{table*}[t]
  \centering
  \begin{adjustbox}{max width=\linewidth}
    \begin{tabular}{l l c c c c c}
        Dataset & Model & 
        \begin{tabular}{c}
            \textsf{DefaultCoT} \\
            Token Count
        \end{tabular} & 
        \begin{tabular}{c}
            \textsf{BeConcise} \\
            Token Count
        \end{tabular} & 
        $T_{\pi}^{*}(A^{*})$ & 
        \begin{tabular}{c}
            \textsf{BeConcise} \\
            Token Reduction
        \end{tabular} & 
        \begin{tabular}{c}
            Upper Bound of \\
            Token Reduction
        \end{tabular} \\
        \hline
        Math-500 & \textsf{GPT-4o} & 635 & 505 & 172 & 1.26x & 3.69x \\
                         & \textsf{GPT-4o-mini} & 611 & 528 & 164 & 1.16x & 3.72x \\
                         & \textsf{Claude-3.5-Sonnet} & 373 & 283 & 105 & 1.32x & 3.56x \\
                         & \textsf{Claude-3.5-Haiku} & 373 & 287 & 143 & 1.30x & 2.61x \\
                         & \textsf{Llama-3.3-70B} & 549 & 475 & 93 & 1.16x & 5.88x \\
        \hline
        GSM8K & \textsf{GPT-4o} & 266 & 190 & 24 & 1.40x & 10.90x \\
                       & \textsf{GPT-4o-mini} & 292 & 216 & 35 & 1.35x & 8.29x \\
                       & \textsf{Claude-3.5-Sonnet} & 200 & 136 & 18 & 1.47x & 11.16x \\
                       & \textsf{Claude-3.5-Haiku} & 212 & 169 & 42 & 1.25x & 5.00x \\
                       & \textsf{Llama-3.3-70B} & 195 & 148 & 27 & 1.32x & 7.24x \\
        \hline
        MMLU-Pro & \textsf{GPT-4o} & 586 & 415 & 121 & 1.41x & 4.83x \\
        Math                  & \textsf{GPT-4o-mini} & 506 & 419 & 132 & 1.21x & 3.85x \\
                          & \textsf{Claude-3.5-Sonnet} & 324 & 227 & 59 & 1.43x & 5.50x \\
                          & \textsf{Claude-3.5-Haiku} & 296 & 237 & 93 & 1.25x & 3.20x \\
                          & \textsf{Llama-3.3-70B} & 552 & 443 & 75 & 1.25x & 7.36x \\
        \hline
    \end{tabular}
\end{adjustbox}
\caption{\label{tab:token-limit}Comparison of token counts on the MMLU-Pro Math, GSM8K, and Math-500 datasets under \textsf{DefaultCoT} and \textsf{BeConcise} along with  $T_{\pi}^{*}(A^{*})$, which is a lower bound on the average token length required to achieve max accuracy. While existing prompt strategies do meaningfully reduce token-length, the lower bound $T_{\pi}^{*}(A^{*})$ illustrates that one can achieve drastically more compression while preserving accuracy.}
\end{table*}

\section{Theoretical Limits of the Length-Performance Tradeoff}
\label{sec:bounds}

Not only does token-complexity provide an interpretable and accurate model of reasoning task performance, it gives insight into how to improve efficiency. Viewing these response-reducing CoT prompts as a form of lossy compression, this raises a natural research question that has so far not been studied: what are the theoretical limits on the optimal tradeoff token-length and accuracy and how far away are existing prompting strategies from this limit? 

We develop a framework to empirically compute bounds on compression performance, inspired by rate-distortion theory. First, we define $\bar{t}_{\pi}(P) \equiv \frac{1}{n}\sum_{i=1}^{n}t(X_{i})$ to be the average token-length under CoT prompt $P$ and LLM $\pi$. We define $\alpha^{*}(T)$ to be the optimal performance for an \emph{average} token budget of $T$ tokens, and $T^{*}(\alpha)$ is the minimum average token length to achieve an accuracy of $\alpha$:
\begin{align}
\alpha^{*}_{\pi}(T) &= \max_{P} \{\text{Acc}_{\pi}(P) : \bar{t}_{\pi}(P)\leq T\} \\
T^{*}_{\pi}(\alpha) &= \min_{P} \{\bar{t}_{\pi}(P) : \text{Acc}_{\pi}(P) \geq \alpha \}
\end{align}
These quantities involve intractable optimizations over CoT prompts $P$. Yet, under the token complexity hypothesis, these optimization problems are greatly simplified, involving only optimization over token-lengths instead of prompts. In fact, the optimization problem is a special case of a knapsack problem, and thus gives rise to a closed-form solution. First, define the empirical CDF of the true token complexities $F_{n}(t) \equiv \frac{1}{n}\sum_{i=1}^{n} \mathbf{1}\{ \tau_{i}^{\pi} \leq t\}$, let $E_{n}(t) \equiv \frac{1}{n}\sum_{i=1}^{n} \tau_{i}^{\pi}\mathbf{1}\{ \tau_{i}^{\pi} \leq t\}$, and let $\bar{\tau}_{\pi}=\frac{1}{N}\sum_{i=1}^{n}\tau_{i}^{\pi}\mathbf{1}\{\tau_{i}^{\pi} < \infty\}$ denote the average token complexity among questions that have finite token complexity.
\begin{theorem} Suppose Assumption~\ref{ass:token-complexity} 
and suppose that for any token counts $\{ t_{i}\}_{i=1}^{n}$ there exists a prompt $P$ such that $t(X_{i}) = t_{i}$. Then,
\begin{align}
\alpha^{*}_{\pi}(T) &= \frac{1}{n}\sum\nolimits_{i=1}^{n} \mathbf{1} \{ \tau^{\pi}_{i} \leq t_{T}\} \\
T^{*}_{\pi}(\alpha) &= \frac{1}{n}\sum\nolimits_{i=1}^{n} \tau^{\pi}_{i}\mathbf{1} \{ \tau^{\pi}_{i} \leq q_{\alpha}\}
\end{align}
where $t_{T} \equiv \inf \{t \in \mathbb{R}: E_{n}(t) = t \}$ and $t_{T} = \infty$ if $t > \bar{\tau}^{\pi}$ and $q_{\alpha}=\sup\{t\in\mathbb{R}:F_{n}(t)\leq\alpha\}$ is the $\alpha$-quantile of the empirical distribution.
\end{theorem}
Intuitively, under Assumption~\ref{ass:token-complexity} the optimal strategy is to only use the minimal number of tokens required to solve the problem, $\tau_{i}^{\pi}$. Under a limited token budget, it is optimal to sort questions by token complexity and solve the questions with shortest token complexity until the budget is filled. Note that this structure is due to the fact that each question is weighted identically, if certain questions had a higher weight than others (e.g. harder questions are more valuable) then the optimal strategy need not have a closed form solution.


We note that it is unlikely that any feasible prompting technique will achieve the upper bound $\alpha^{*}_{\pi}(T)$ or the lower bound $T^{*}(\alpha)$, as doing so requires (1) knowing the token complexities, (2) inducing a chain-of-thought that exactly matches the token complexity, (3) prioritizing easier questions over harder ones. 

Nevertheless $\alpha^{*}_{\pi}(T)$ provides a computable upper bound on maximum accuracy for a particular token budget, which we plot in Figure~\ref{fig:tradeoff} under the label `oracle upper bound', using estimated token complexities $\hat{\tau}_{i}^{\pi}$. Across all the LLMs and benchmarks we consider, we find that this is indeed serves as an upper bound on the performance of the prompting strategies we consider across response lengths, especially for more difficult datasets such as MATH-500 and MMLU-Pro Math. Yet, for GSM8K in Appendix~\ref{sec:appendix_tradeoff}, we see that this gap with the upper bound $\alpha^{*}_{\pi}(T)$ is much smaller, illustrating that while $\alpha^{*}_{\pi}(T)$ may be challenging to achieve it still gives a reasonable upper bound on performance.

Finally, we circle back to the observation made in Section~\ref{sec:tradeoff} that one can substantially reduce the length of the chain-of-thought (with prompts like \textsf{BeConcise} or \textsf{NoProperGrammar}) while maintaining a similar accuracy to \textsf{DefaultCoT}. This leads to a natural question: what is the lower bound on the number of tokens needed in order to achieve the best possible accuracy? Under the token-complexity hypothesis, we obtain a simple closed-form expression for this lower bound.

\begin{corollary}
\label{cor:lossless}
Suppose Assumption 1 holds. Let  $A^{*} = \frac{1}{n}\sum_{i=1}^{n} \mathbf{1}\{\tau_{i}^{\pi} < \infty\}$ be the maximum possible accuracy achieved by the LLM on the dataset. The number of tokens required to achieve accuracy $\alpha = A^{*}$ under the optimal token allocation:
\begin{align}
T_{\pi}^{*}(A^{*}) &=\bar{\tau}_{\pi}= \frac{1}{n}\sum_{i=1}^{n}\tau_{i}^{\pi}\mathbf{1}\{\tau_{i}^{\pi} < \infty\}
\end{align}
\end{corollary}
Corollary~\ref{cor:lossless} gives the number of tokens required for `lossless compression', i.e. achieving the best possible accuracy. 
Remarkably, this lower bound is only the \emph{mean} of the token-complexities, which can be much smaller than the average token-length of existing prompting strategies. In Table~\ref{tab:token-limit}, we consider the MMLU-Pro Math dataset across several LLMs, and we compare the lower bound $T_{\pi}^{*}(A^{*})$ with the token counts of \textsf{DefaultCoT} and \textsf{BeConcise}. Across all the LLMs, while \textsf{BeConcise} reduces the token counts by 1.2-1.4x, the token reduction achieved by the optimal compression scheme is \textbf{3.2-11.2x}, and is especially high for GSM8K. Along with the results in Figure~\ref{fig:tradeoff}, this illustrates that while the upper bound may not be exactly attainable, there may be significant room for improvement especially for easier datasets.

Finally, we note that token complexity may be of broader interest as a measure of reasoning capabilities, even for benchmarks which are `saturated'. For instance, both Claude 3.5 Sonnet and Claude 3.5 Haiku achieve a similar accuracy on GMS8K (97\% vs 95\% under \textsf{DefaultCoT}). Yet the average token complexity $\bar{\tau}_{\pi}$ is 42.4 for Haiku and 17.9 for Sonnet, showing that even if the accuracy is similar, Sonnet can achieve it with much fewer tokens.

\section{Towards the Theoretical Limit}

The token complexity hypothesis not only provides limits on the efficiency of an optimal chain-of-thought compression scheme, it highlights the importance of \emph{adaptive} compression -- using shorter chains-of-thought for easier questions -- for approaching these limits. This provides theoretical motivation for methods developed in recent works (e.g.~\cite{han2024token,kang2024c3ot}) designed to calibrate the length of the chain-of-thought to problem difficulty. In this section, we illustrate how our empirical framework can help evaluate and contextualize recent advances in adaptive reasoning.


\begin{figure}
  \includegraphics[width=0.5\linewidth]{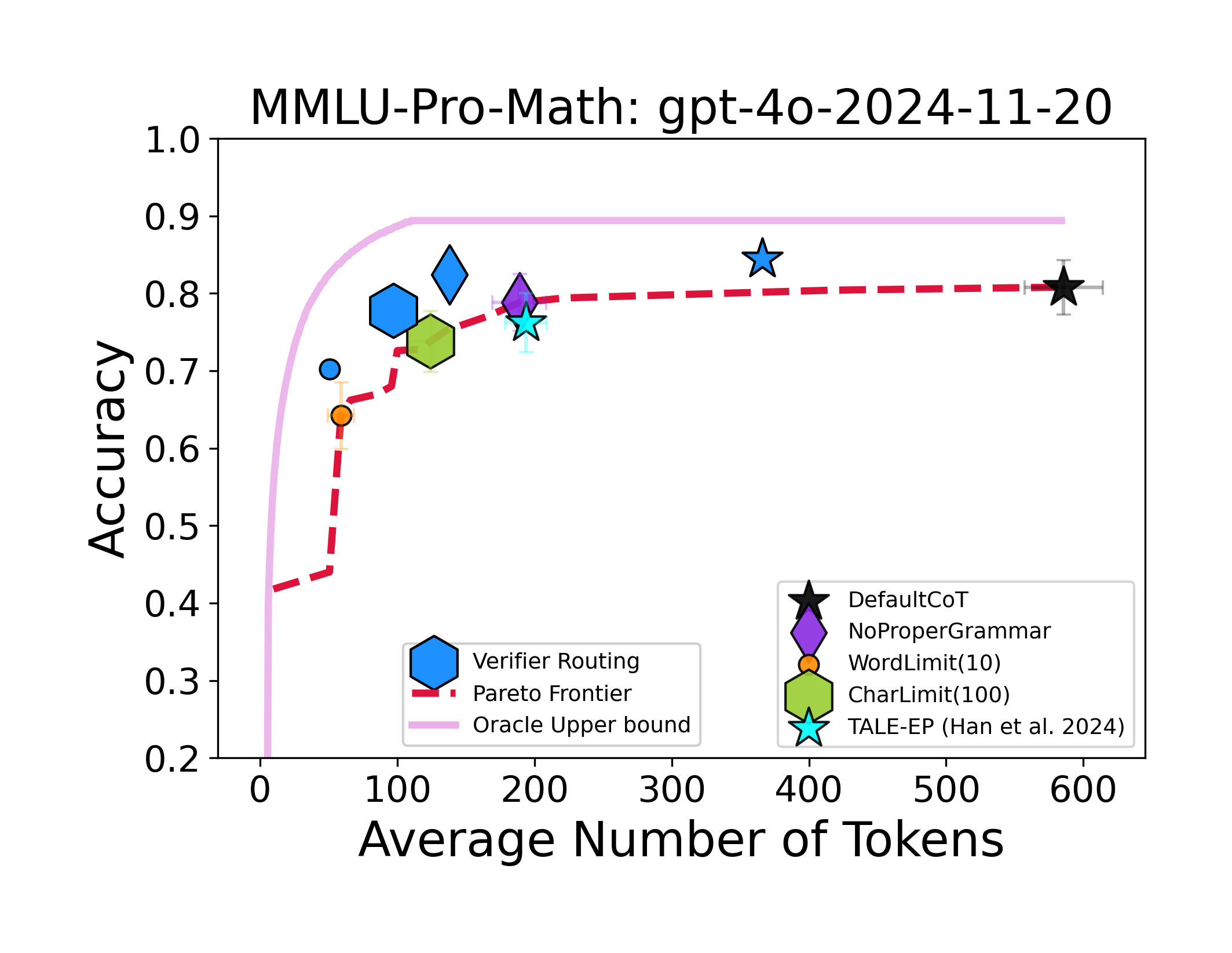}
  \includegraphics[width=0.5\linewidth]{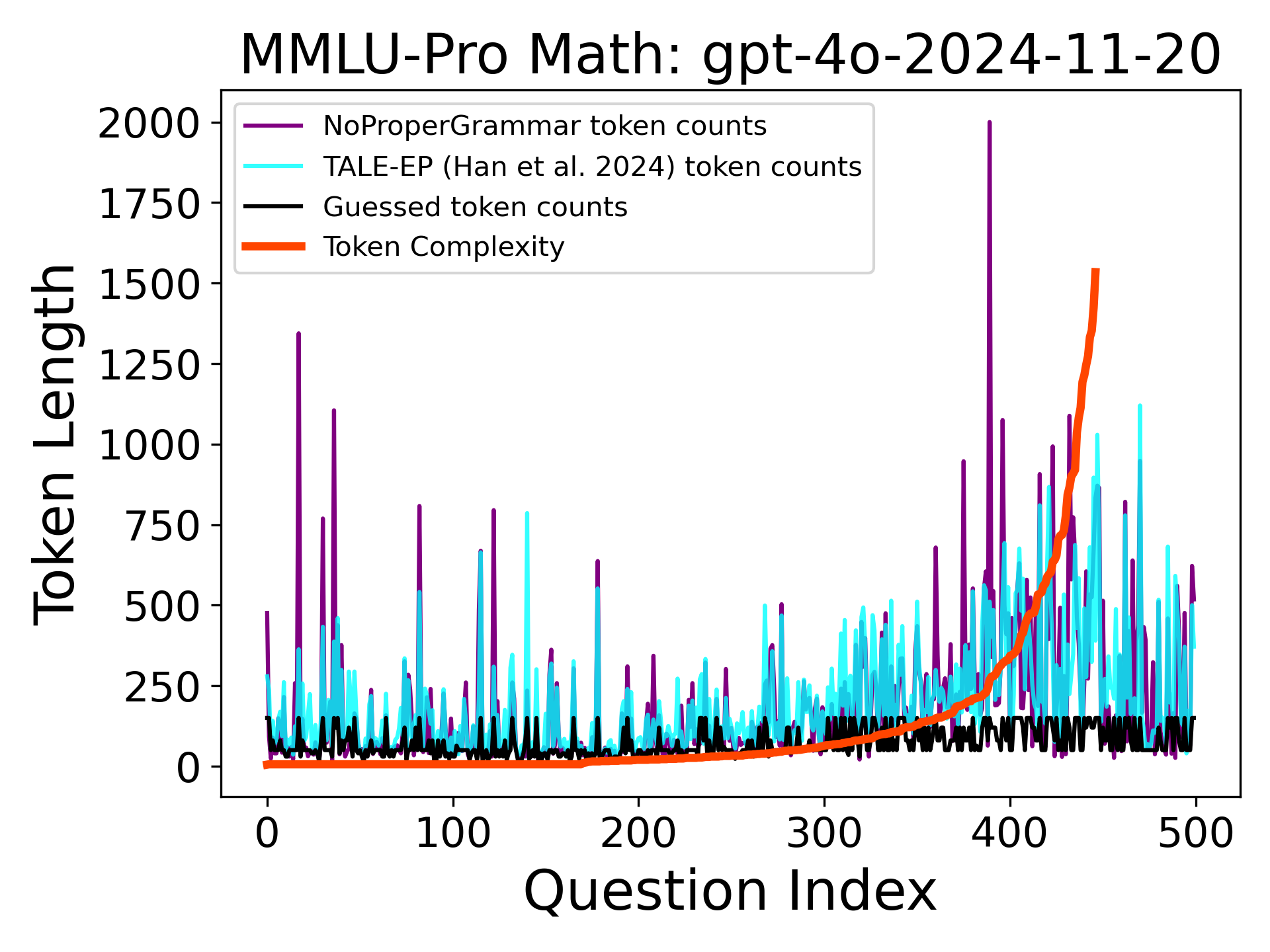}
  \caption {\label{fig:routing}Does adaptive compression improve the accuracy-length tradeoff? (Left) Average token length vs Accuracy for GPT-4o on MMLU-Pro-Math under different prompt routing strategies. TALE-EP~\citep{han2024token} (light green), which first has the LLM guess the number of tokens to use, is slightly below the Pareto curve achieved by simpler prompting strategies. 
  Only verifier-based routing (in blue) is able to achieve a better tradeoff, closer to the upper bound. (Right) We compare token counts with token complexities across questions in MMLU-Pro Math, which are sorted by token complexity. The correlation between the token count of TALE-EP and the true token-complexities (Spearman $\rho = 0.51$) is similar to the correlation observed for \textsf{NoProperGrammar} (Spearman $\rho = 0.54$), which does not explicitly prompt the LLM to reason adaptively. This can help explain why both methods have similar accuracy-length tradeoffs.}
\end{figure}

As a proof of concept, we consider two prompting strategies designed to adjust reasoning effort to problem difficulty. \cite{han2024token} propose a two-step procedure called TALE-EP, which first (1) prompts the LLM to guess the minimum tokens required to solve the question and then (2) prompts the LLM to think step-by-step using the guessed number of tokens. Using our experimental results, we can compare whether adaptively adjusting the reasoning effort improves upon the accuracy-length tradeoff. The left panel of Figure~\ref{fig:routing} plots accuracy and average token-length of TALE-EP for GPT-4o on MMLU-Pro Math compared to several prompts from our experiments. Surprisingly, we observe that TALE-EP is slightly below the trade-off curve from our simple prompting strategies. More specifically, it produces a slightly worse accuracy than \textsf{NoProperGrammar}, even though it uses more tokens on average. This is consistent across other benchmarks (see Appendix~\ref{sec:appendix-routing}). 

We find a potential explanation in the right panel of Figure~\ref{fig:routing}, which plots the token lengths used by TALE-EP and \textsf{NoProperGrammar} along with the token complexity. We find that that the token counts produced by \textsf{NoProperGrammar} have a surprisingly high correlation with the true token complexity, which is equivalent to the correlation for the token counts of TALE-EP (Spearman $\rho = 0.51$ for TALE-EP and $\rho = 0.54$ for \textsf{NoProperGrammar},  see Appendix~\ref{sec:appendix_tables} for other prompts). Thus, this shows that LLMs \emph{natively} adjust response length to the difficulty of the problem, even without explicitly being prompted to do so. And moreover, this capability is \emph{equivalent} to more sophisticated prompting strategies, demonstrating that the simple prompting strategies we test are surprisingly strong baselines and that LLMs may struggle to estimate token complexity accurately. Nonetheless, methods based on fine-tuning (e.g.~\cite{kang2024c3ot} or TALE-PT in~\cite{han2024token}) may be able to outperform these simple prompting strategies, which we leave to future research.

This result suggests that it may be challenging to achieve a better accuracy-length tradeoff. Nonetheless, if one has access to a perfect verifier~\citep{brown2024large, setlur2024rewarding}, we can substantially improve the tradeoff. \textsf{Verifier Routing} first (1) prompts the LLM to without chain-of-thought (\textsf{NoCoT}) to obtain an initial solution and (2) if the solution is incorrect it uses a longer chain-of-thought prompt (e.g. \textsf{DefaultCoT}) to produce a better answer. The blue dots in the left panel Figure~\ref{fig:routing} show performance of \textsf{Verifier Routing} with a selection of four longer prompts. This routing strategy achieves a significantly better trade-off between reasoning length and performance, and approaches the theoretical upper bound. 
Altogether, this shows that the upper bound $\alpha_{\pi}^{*}(T)$ can be approached through adaptive compression, although this requires a very accurate signal of problem difficulty and motivates further research.



\section{Conclusion}

Our study presents a systematic investigation into the trade-off between reasoning length and performance in large language models (LLMs), across different prompts. We demonstrate that this trade-off follows a universal Pareto curve, suggesting that reasoning length, rather than specific compression strategies, primarily determines accuracy. Introducing the concept of token complexity, we find that LLM performance at the question-level exhibits a sharp threshold behavior -- it is successful only if its chain-of-thought is above a token threshold. Our analysis, inspired by rate-distortion theory, reveals that existing prompt-based compression strategies operate far from the optimal accuracy-length frontier, highlighting substantial room for improvement. Our work enables researchers to contextualize the performance of new methodologies for improving chain-of-thought compression and assess adaptivity of LLM reasoning.




\bibliography{custom}
\bibliographystyle{colm2025_conference}

\appendix
\section{Limitations}

While our study provides novel insights into the relationship between reasoning length and LLM performance, there are several limitations. First, our study is limited to mathematical reasoning tasks, and it remains to be seen whether similar accuracy-length tradeoffs and token complexity thresholds apply to other domains such as commonsense reasoning, code generation, or open-ended text generation. The theoretical upper bound on accuracy-length tradeoff is rarely achieved due to the practical challenges in estimating token complexities and generating precisely compressed responses. Computing token complexity is computationally expensive, requiring multiple generations per question, making it impractical for large-scale applications. Our approach also assumes that token complexity is well-defined and consistent across different models and tasks, which may not hold for all LLMs or highly complex benchmarks. Furthermore, our experiments were limited to a fixed set of 31 prompts, and exploring a broader range of compression strategies, including model fine-tuning or iterative refinement, could potentially expand the compression frontier. Finally, our analysis is most relevant for strong models on moderately difficult benchmarks; weaker models or extremely challenging tasks may exhibit different behavior.

\section{Correlation with Token Complexity}
\label{sec:appendix_tables}

The following is a table describing the Spearman correlation between token-lengths with token-complexity across different prompts for GPT-4o on MMLU-Pro Math. While the ordering of which prompts have the highest correlation changes across different models and benchmarks, we observe widely that the correlation ranges between 0 - 0.6.

\begin{table}[h]
    \centering
    \begin{tabular}{l c}
        \hline
        Prompt ID & \begin{tabular}{c}
        Spearman $\rho$
        \end{tabular}\\
        \hline
        \textsf{CharLimit(50)} & 0.57 \\
        \textsf{NoSpaces} & 0.56 \\
        \textsf{OnlyNumbers} & 0.55 \\
        \textsf{CharLimit(100)} & 0.55 \\
        \textsf{NoProperGrammar} & 0.54 \\
        \vdots & \vdots \\
        \textsf{TokenLimit(10)} & 0.38 \\
        \textsf{WordLimit(15)} & 0.37 \\
        \textsf{WordLimit(5)} & 0.29 \\
        \textsf{WordLimit(10)} & 0.28 \\
        \textsf{NoCoT} & 0.08 \\
        \hline
    \end{tabular}
    \caption{\label{tab:kendall}GPT-4o on MMLU-Pro Math: Spearman correlation of token-lengths with token-complexity across different prompts.}
    \label{tab:kendall_corr}
\end{table}

\begin{figure*}
\centering
  \includegraphics[width=0.5\linewidth]{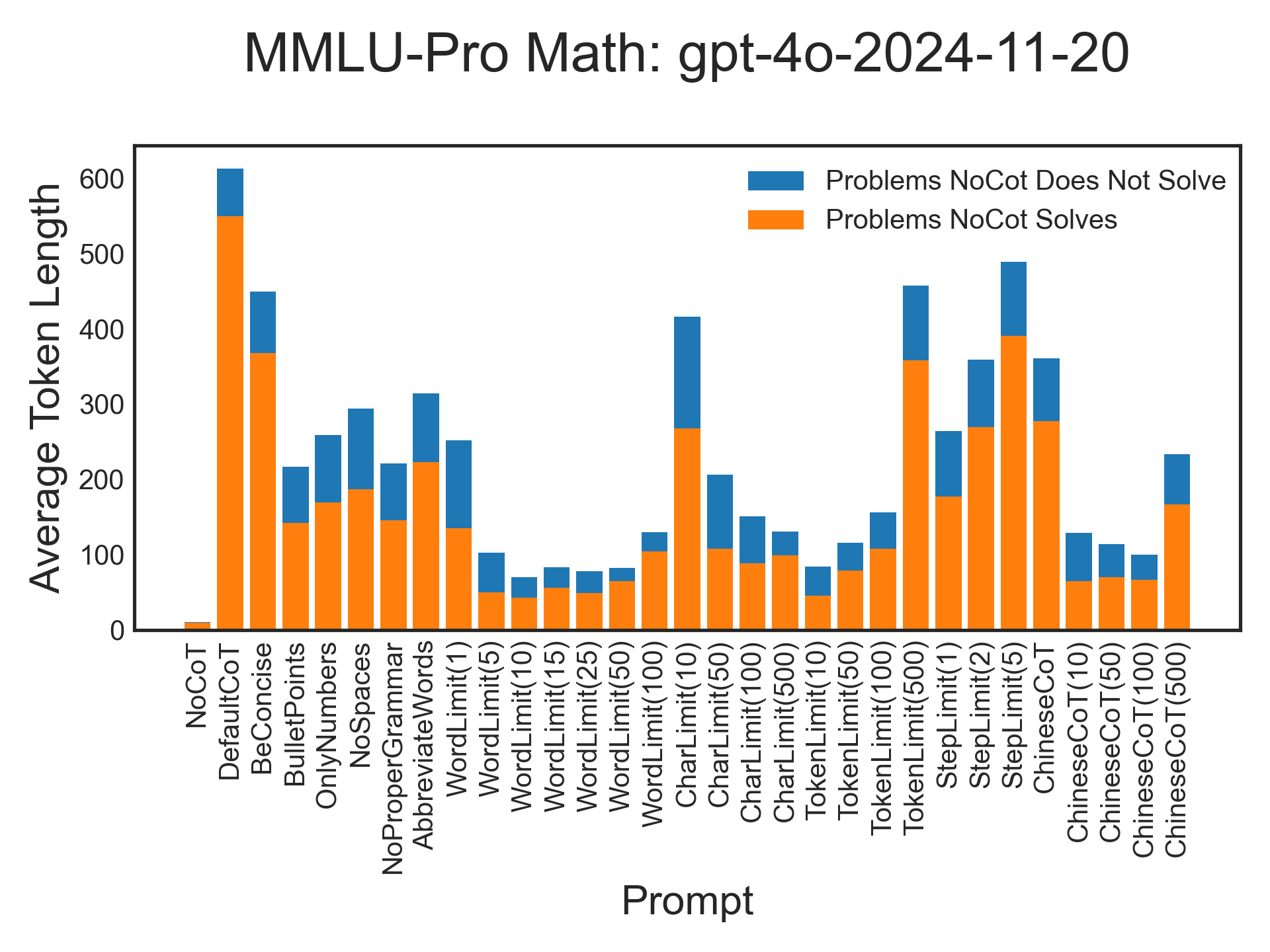}

  \caption {\label{fig:adaptive} Do LLM's tailor response length to problem difficulty? Average token length produced by GPT-4o on MMLU-Pro Math across prompts in Table~\ref{tab:prompts}, split by problems which can be solved without chain-of-thought and problems which \text{NoCoT} does not successfully solve. Response lengths are consistently higher for the problems \text{NoCoT} unsuccessfully solves.}
\end{figure*}

In Figure~\ref{fig:adaptive}, we consider the average token lengths of GPT-4o on MMLU-Pro Math. We split the problems into two categories: problems that the LLM successfully solves without chain-of-thought (\textsf{NoCot}), and the problems  \textsf{NoCot} unsuccessfully solves. Intuitively, the first class of problems are `easy', since they do not require any chain-of-thought, and the rest are harder.  Across all the prompts we consider, we observe that the average token-length among `easy' problems is universally smaller than among `harder' problems. Surprisingly, this is even true for the \textsf{WordLimit}/\textsf{TokenLimit}/\textsf{CharLimit}/etc. prompts, which are supposed to limit the LLM's response to a fixed length. Nevertheless, despite the evidence of adaptive response lengths, this also shows that there is a lot of room for improvement: even though the LLM can solve the problem without any chain-of-thought, the model still proceeds to generate long chains-of-thought.

\section{Accuracy and Average Token Length}

Below are tables which describe the accuracy and average token length for several prompts across models and benchmarks.

\begin{table}
\centering
\begin{adjustbox}{max width=0.5\columnwidth}
\centering
\begin{tabular}{l | l | l | l | l}
\hline
Model & Prompt & Accuracy & Token Length \\
\hline

\multirow{8}{*}{\textsf{GPT-4o}}
  & NoCoT & 41.8 & 9 \\
 & DefaultCoT & 80.8 & 585 \\
 & BeConcise & 80.4 & 415 \\
 & BulletPoints & 75.0 & 185 \\
 & OnlyNumbers & 79.4 & 221 \\
 & NoSpaces & 78.8 & 248 \\
 & NoProperGrammar & 78.8 & 189 \\
 & AbbreviateWords & 74.2 & 275 \\
 \hline
\multirow{8}{*}{\textsf{GPT-4o Mini}} & NoCoT & 32.4 & 6 \\
 & DefaultCoT & 75.0 & 506 \\
 & BeConcise & 74.8 & 418 \\
 & BulletPoints & 63.0 & 128 \\
 & OnlyNumbers & 68.8 & 189 \\
 & NoSpaces & 66.2 & 219 \\
 & NoProperGrammar & 61.0 & 119 \\
 & AbbreviateWords & 56.6 & 259 \\
 \hline
\multirow{8}{*}{\textsf{Claude 3.5 Sonnet}} & NoCoT & 48.2 & 7 \\
 & DefaultCoT & 80.8 & 324 \\
 & BeConcise & 78.2 & 227 \\
 & BulletPoints & 75.8 & 167 \\
 & OnlyNumbers & 76.6 & 130 \\
 & NoSpaces & 76.8 & 187 \\
 & NoProperGrammar & 76.8 & 172 \\
 & AbbreviateWords & 77.2 & 220 \\
 \hline
\multirow{8}{*}{\textsf{Claude 3.5 Haiku}} & NoCoT & 30.2 & 19 \\
 & DefaultCoT & 68.8 & 296 \\
 & BeConcise & 68.4 & 237 \\
 & BulletPoints & 61.8 & 155 \\
 & OnlyNumbers & 63.2 & 147 \\
 & NoSpaces & 62.6 & 193 \\
 & NoProperGrammar & 63.2 & 169 \\
 & AbbreviateWords & 62.6 & 243 \\
 \hline
\multirow{8}{*}{\textsf{LLaMA 3.3 70B}} & NoCoT & 42.6 & 20 \\
 & DefaultCoT & 74.4 & 551 \\
 & BeConcise & 76.6 & 442 \\
 & BulletPoints & 70.0 & 193 \\
 & OnlyNumbers & 63.4 & 304 \\
 & NoSpaces & 67.2 & 330 \\
 & NoProperGrammar & 68.4 & 217 \\
 & AbbreviateWords & 70.2 & 327 \\
\hline
\end{tabular}
\end{adjustbox}
\caption{\label{tab:MMLU_perf}Accuracy-length tradeoff on MMLU-Pro Math}
\end{table}

\begin{table}
\centering
\begin{adjustbox}{max width=0.5\columnwidth}
\begin{tabular}{l | l | l | l | l}
\hline
Model & Prompt & Accuracy & Token Length \\
\hline

\multirow{8}{*}{\textsf{GPT-4o}}
& NoCoT & 70.0 & 30 \\
& DefaultCoT & 92.8 & 266 \\
& BeConcise & 96.6 & 190 \\
 & BulletPoints & 96.6 & 104 \\
 & OnlyNumbers & 97.4 & 76 \\
 & NoSpaces & 96.4 & 93 \\
 & NoProperGrammar & 95.8 & 71 \\
 & AbbreviateWords & 95.0 & 97 \\
 \hline
\multirow{8}{*}{\textsf{GPT-4o Mini}} 
& NoCoT & 28.0 & 6 \\
 & DefaultCoT & 94.6 & 292 \\
 & BeConcise & 94.2 & 216 \\
 & BulletPoints & 92.2 & 97 \\
 & OnlyNumbers & 92.0 & 77 \\
 & NoSpaces & 86.6 & 62 \\
 & NoProperGrammar & 91.0 & 76 \\
 & AbbreviateWords & 72.8 & 121 \\
 \hline
\multirow{8}{*}{\textsf{Claude 3.5 Sonnet}} 
& NoCoT & 67.8 & 7 \\
 & DefaultCoT & 97.0 & 200 \\
 & BeConcise & 97.4 & 136 \\
 & BulletPoints & 97.0 & 100 \\
 & OnlyNumbers & 96.0 & 66 \\
 & NoSpaces & 97.0 & 111 \\
 & NoProperGrammar & 97.8 & 111 \\
 & AbbreviateWords & 95.0 & 119 \\
 \hline
\multirow{8}{*}{\textsf{Claude 3.5 Haiku}} 
 & NoCoT & 30.6 & 8 \\
 & DefaultCoT & 95.2 & 211 \\
 & BeConcise & 94.4 & 169 \\
 & BulletPoints & 92.8 & 114 \\
 & OnlyNumbers & 93.0 & 87 \\
 & NoSpaces & 92.4 & 116 \\
 & NoProperGrammar & 91.8 & 120 \\
 & AbbreviateWords & 90.8 & 137 \\
 \hline
\multirow{8}{*}{\textsf{LLaMA 3.3 70B}} 
 & NoCoT & 88.6 & 98 \\
 & DefaultCoT & 96.2 & 194 \\
 & BeConcise & 95.8 & 147 \\
 & BulletPoints & 96.2 & 80 \\
 & OnlyNumbers & 89.6 & 81 \\
 & NoSpaces & 92.2 & 102 \\
 & NoProperGrammar & 95.4 & 91 \\
 & AbbreviateWords & 94.0 & 136 \\
\hline
\end{tabular}
\end{adjustbox}
\caption{\label{tab:MMLU_perf}Accuracy-length tradeoff on GSM8K}
\end{table}

\begin{table}
\centering
\begin{adjustbox}{max width=0.5\columnwidth}
\begin{tabular}{l | l | l | l | l}
\hline
Model & Prompt & Accuracy & Token Length \\
\hline

\multirow{8}{*}{\textsf{GPT-4o}}
& NoCoT & 55.6 & 116 \\
& DefaultCoT & 72.8 & 634 \\
& BeConcise & 71.6 & 505 \\
& BulletPoints & 70.6 & 302 \\
& OnlyNumbers & 68.4 & 272 \\
& NoSpaces & 71.2 & 368 \\
& NoProperGrammar & 68.8 & 286 \\
& AbbreviateWords & 72.0 & 416 \\
 \hline
\multirow{8}{*}{\textsf{GPT-4o Mini}} 
& NoCoT & 25.6 & 9 \\
& DefaultCoT & 70.4 & 610 \\
& BeConcise & 72.0 & 528 \\
& BulletPoints & 68.8 & 266 \\
& OnlyNumbers & 66.2 & 306 \\
& NoSpaces & 67.6 & 390 \\
& NoProperGrammar & 66.8 & 254 \\
& AbbreviateWords & 62.0 & 367 \\
 \hline
\multirow{8}{*}{\textsf{Claude 3.5 Sonnet}} 
& NoCoT & 39.0 & 9 \\
& DefaultCoT & 74.8 & 373 \\
& BeConcise & 73.0 & 282 \\
& BulletPoints & 70.0 & 203 \\
& OnlyNumbers & 63.8 & 172 \\
& NoSpaces & 70.4 & 240 \\
& NoProperGrammar & 69.6 & 225 \\
& AbbreviateWords & 71.8 & 263 \\
 \hline
\multirow{8}{*}{\textsf{Claude 3.5 Haiku}} 
& NoCoT & 25.8 & 19 \\
& DefaultCoT & 66.0 & 373 \\
& BeConcise & 64.0 & 286 \\
& BulletPoints & 60.0 & 243 \\
& OnlyNumbers & 59.0 & 193 \\
& NoSpaces & 56.4 & 257 \\
& NoProperGrammar & 61.4 & 253 \\
& AbbreviateWords & 59.4 & 294 \\
 \hline
\multirow{8}{*}{\textsf{LLaMA 3.3 70B}} 
& NoCoT & 33.8 & 50 \\
& DefaultCoT & 55.4 & 549 \\
& BeConcise & 67.0 & 475 \\
& BulletPoints & 63.6 & 238 \\
& OnlyNumbers & 57.0 & 362 \\
& NoSpaces & 57.2 & 439 \\
& NoProperGrammar & 61.2 & 294 \\
& AbbreviateWords & 63.8 & 383 \\
\hline
\end{tabular}
\end{adjustbox}
\caption{\label{tab:MMLU_perf}Accuracy-length tradeoff on MATH-500}
\end{table}

\section{Proof of Theorem 1}

Under Assumption~\ref{ass:token-complexity}, we have that the accuracy can be represented as
\begin{align}
    \text{Acc}_{\pi}(P) = \frac{1}{n}\sum_{i=1}^{n} \mathbf{1}\{\tau_{i}\geq t_{i}\}
\end{align}
where we suppress dependence on $\pi$ for now. The optimization problem then becomes
\begin{align}
\alpha^{*}_{\pi}(T) =  \max_{t} & \frac{1}{n}\sum_{i=1}^{n} \mathbf{1}\{\tau_{i}\geq t_{i}\} \\
\text{s.t. } & \frac{1}{n} \sum_{i=1}^{n} t_{i} \leq T
\end{align}
Note that the optimal strategy is either to set $t_{i} = \tau_{i}$ or $0$. Thus, this is equivalent to a Knapsack problem,
\begin{align}
\alpha^{*}_{\pi}(T) =  \max_{x_{i} \in \{0,1\}} & \frac{1}{n}\sum_{i=1}^{n} x_{i} \\
\text{s.t. } & \frac{1}{n} \sum_{i=1}^{n} \tau_{i}x_{i} \leq T
\end{align}
where $x_{i}$ are indicators whether the LLM tries to solve the problem or not. This is a special case where there are unit rewards (since all problems are weighted equally). The optimal strategy in this case is a greedy policy, sorting the questions in increasing order of token complexity $\tau_{(1)} < ... < \tau_{(n)}$ and only solving as many as can fit within budget. 
$t_{T} \equiv \inf \{t\in \mathbb{R}: E_{n}(t) = t\}$ is the highest value of $\tau$ that the LLM puts into the knapsack after ordering them greedily, so thus $\alpha_{n}^{*}(T)$ is the total number of questions with token complexity less than $\tau$. The proof for $T^{*}(\alpha)$ proceeds similarly, as the optimal strategy is identical.

\section{Example Prompt}

We use the following template for our prompts:

\fbox{\begin{minipage}{\linewidth}
Answer the following question. {PROMPT}
Question: {QUESTION}
The last line of your response 
should be of the following format: 
'Answer: ANSWER' (without quotes) 
where ANSWER is your final answer.
\end{minipage}
}

\section{Tradeoff curves for more models and benchmarks}
\label{sec:appendix_tradeoff}

In this section, we present results for the performance of different models and benchmarks (GSM8K and MATH-500). We see broadly that performance across all prompts lies on the same trade-off curve.


\begin{figure}
  \centering
  \includegraphics[width=0.48\linewidth]{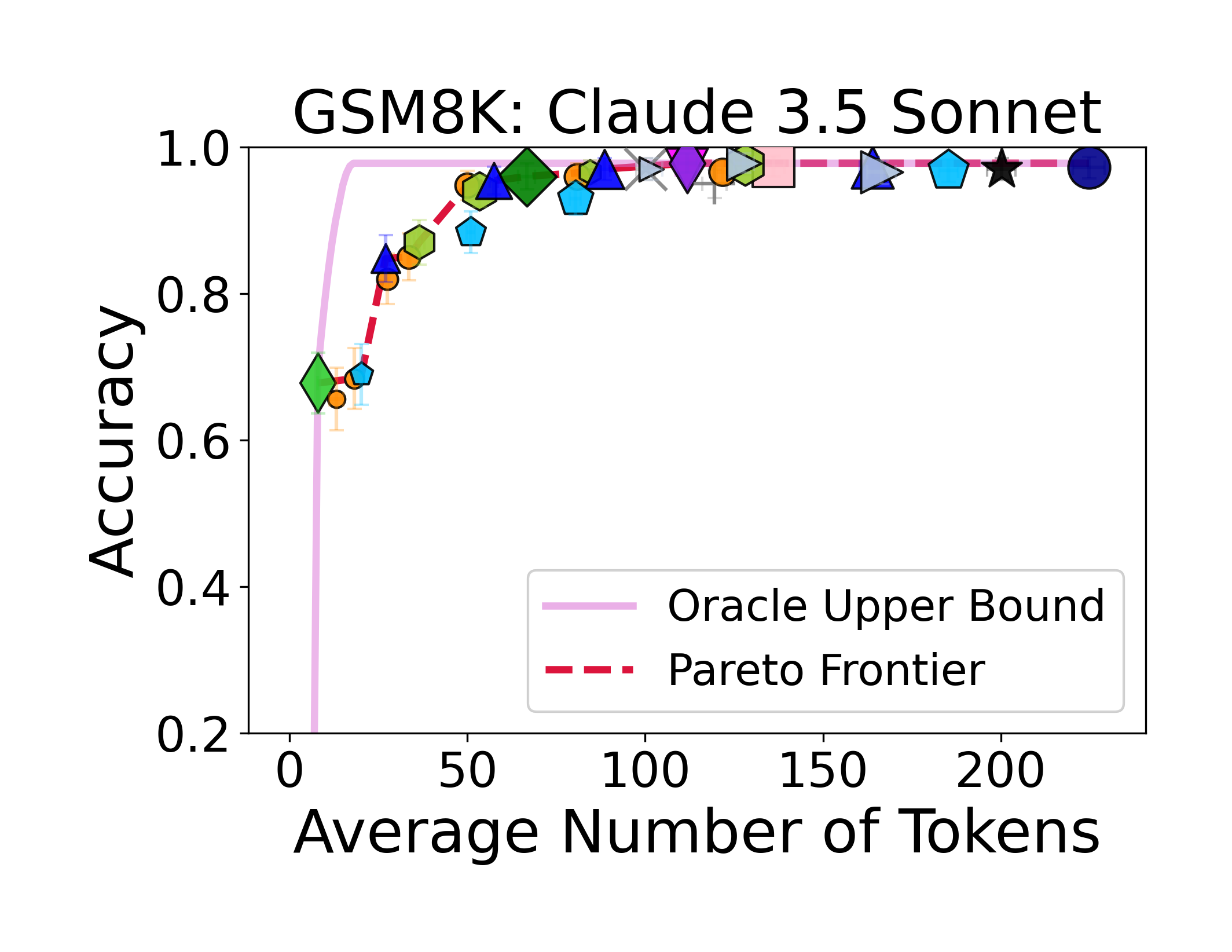}
  \includegraphics[width=0.48\linewidth]{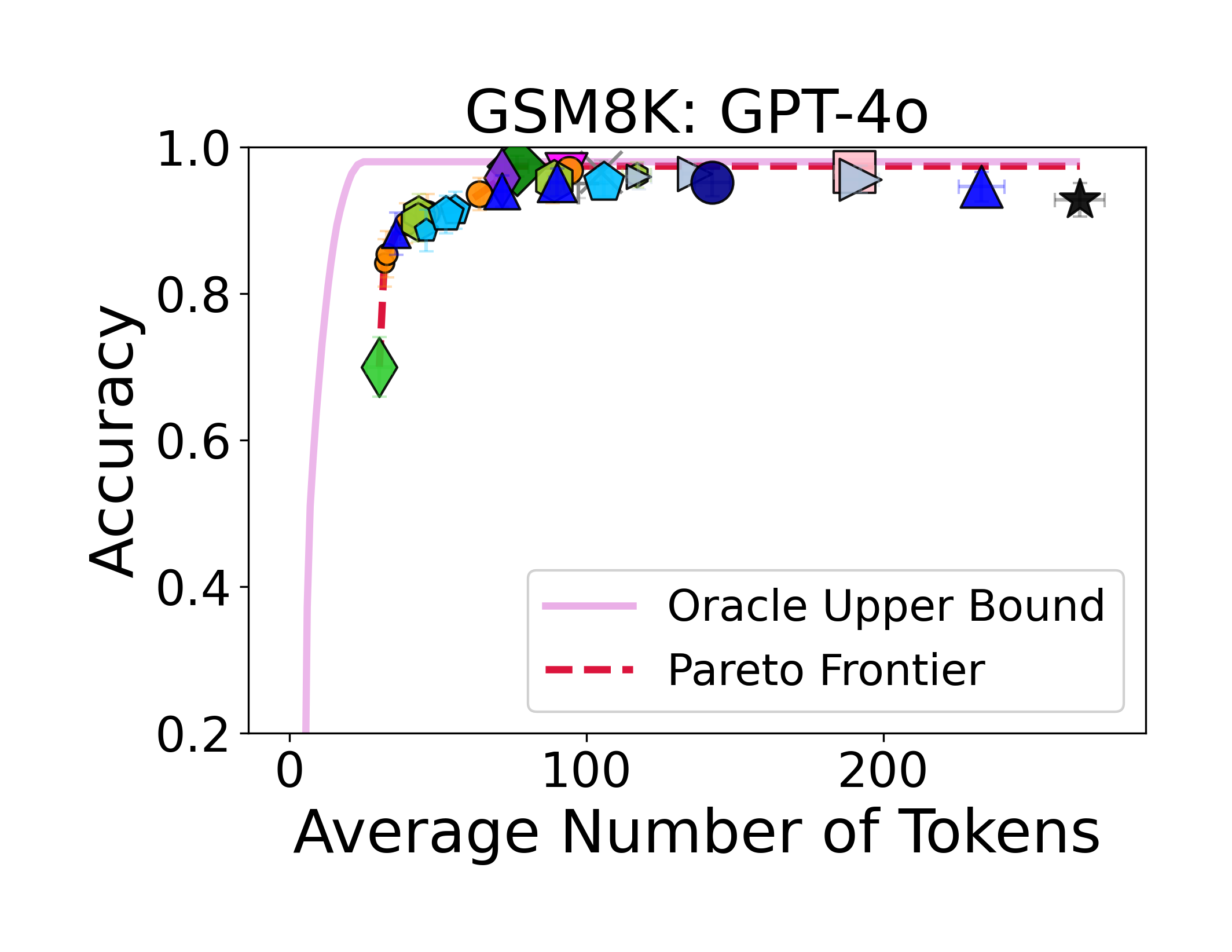}
  
  \includegraphics[width=0.48\linewidth]{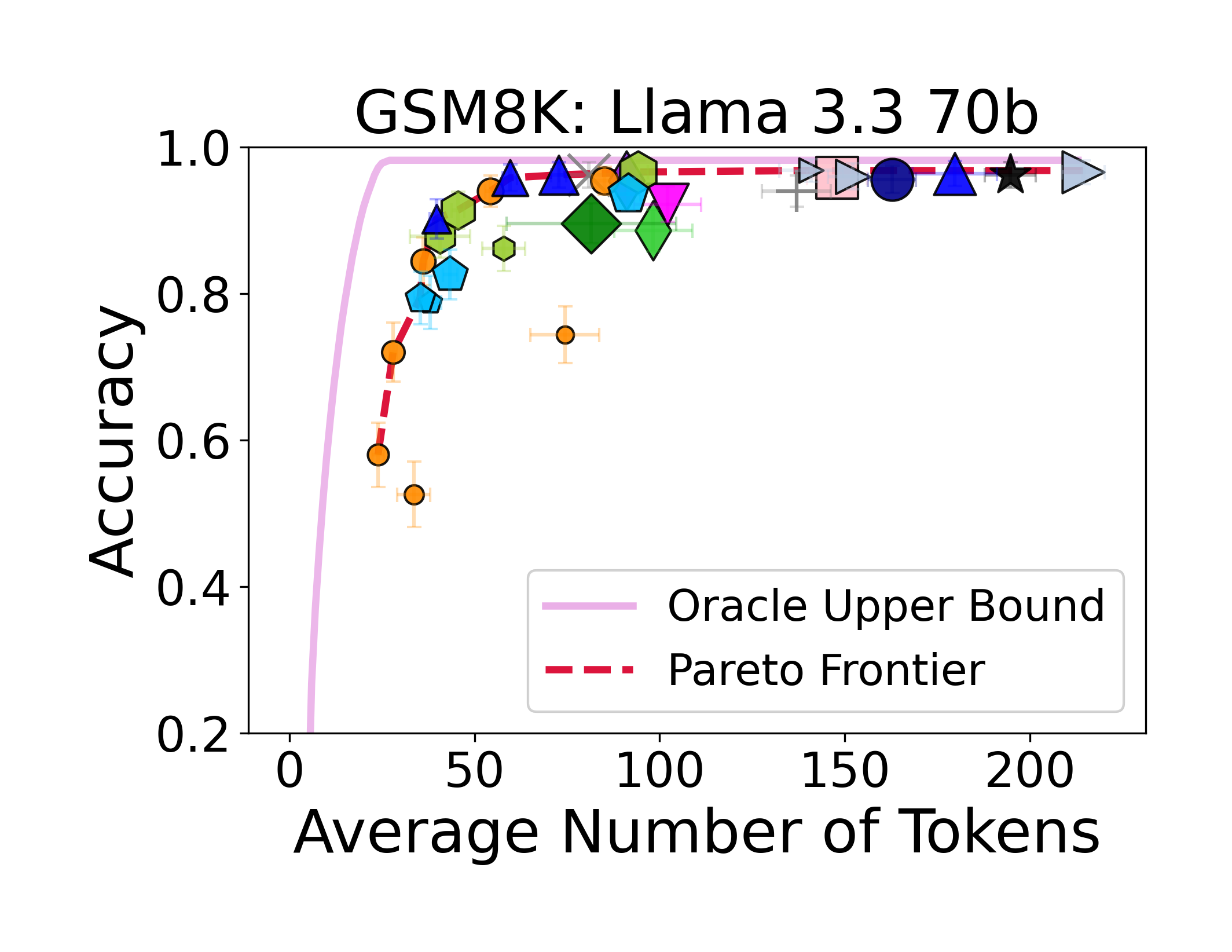}
  \includegraphics[width=0.48\linewidth]{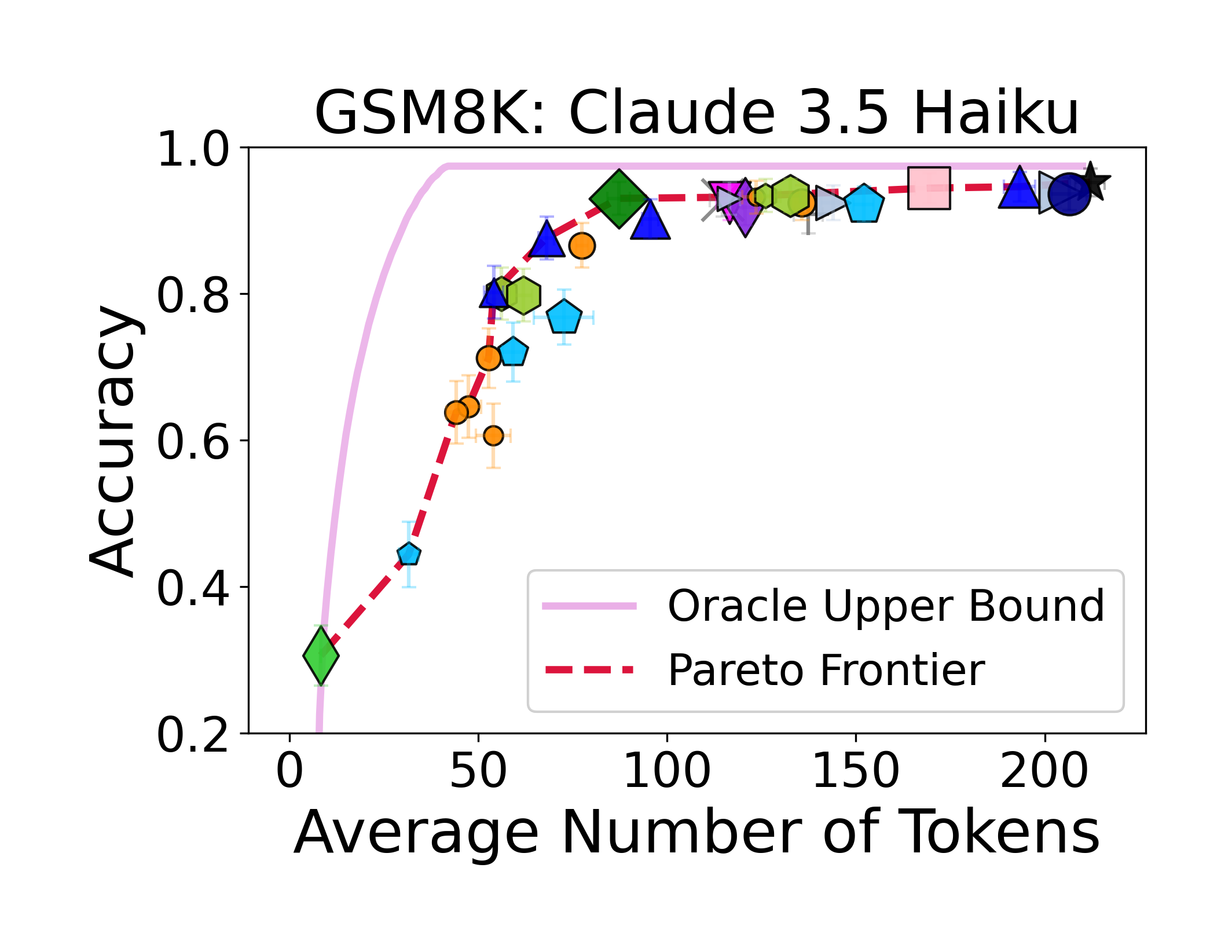}

\includegraphics[width=0.48\linewidth]{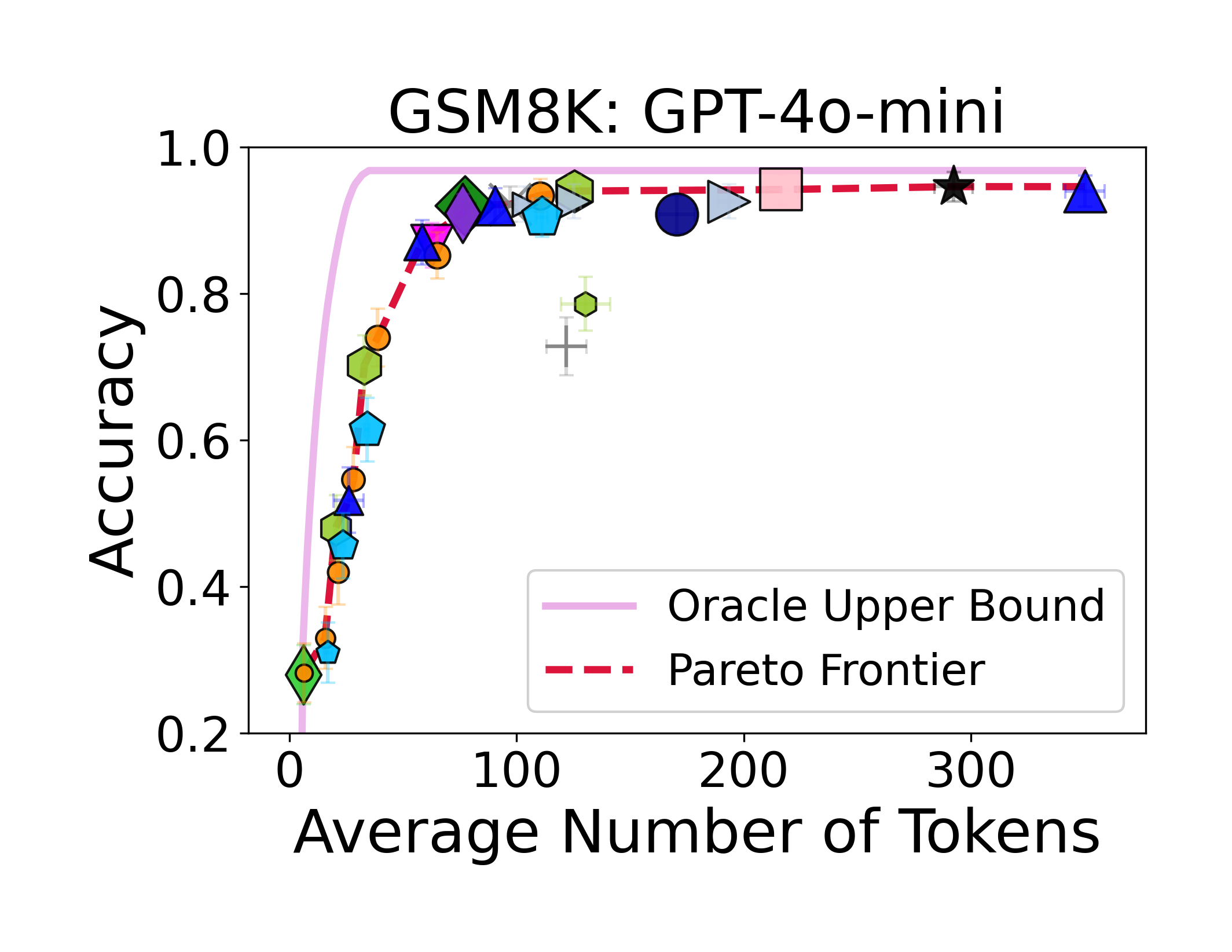}
    
  \captionsetup{aboveskip=1pt, belowskip=2pt}
  \caption{Tradeoff Curves for GSM8K}
  \label{fig:GSM8K_ClaudePareto}
  \end{figure}

\begin{figure}
  \centering
  \includegraphics[width=0.48\linewidth]{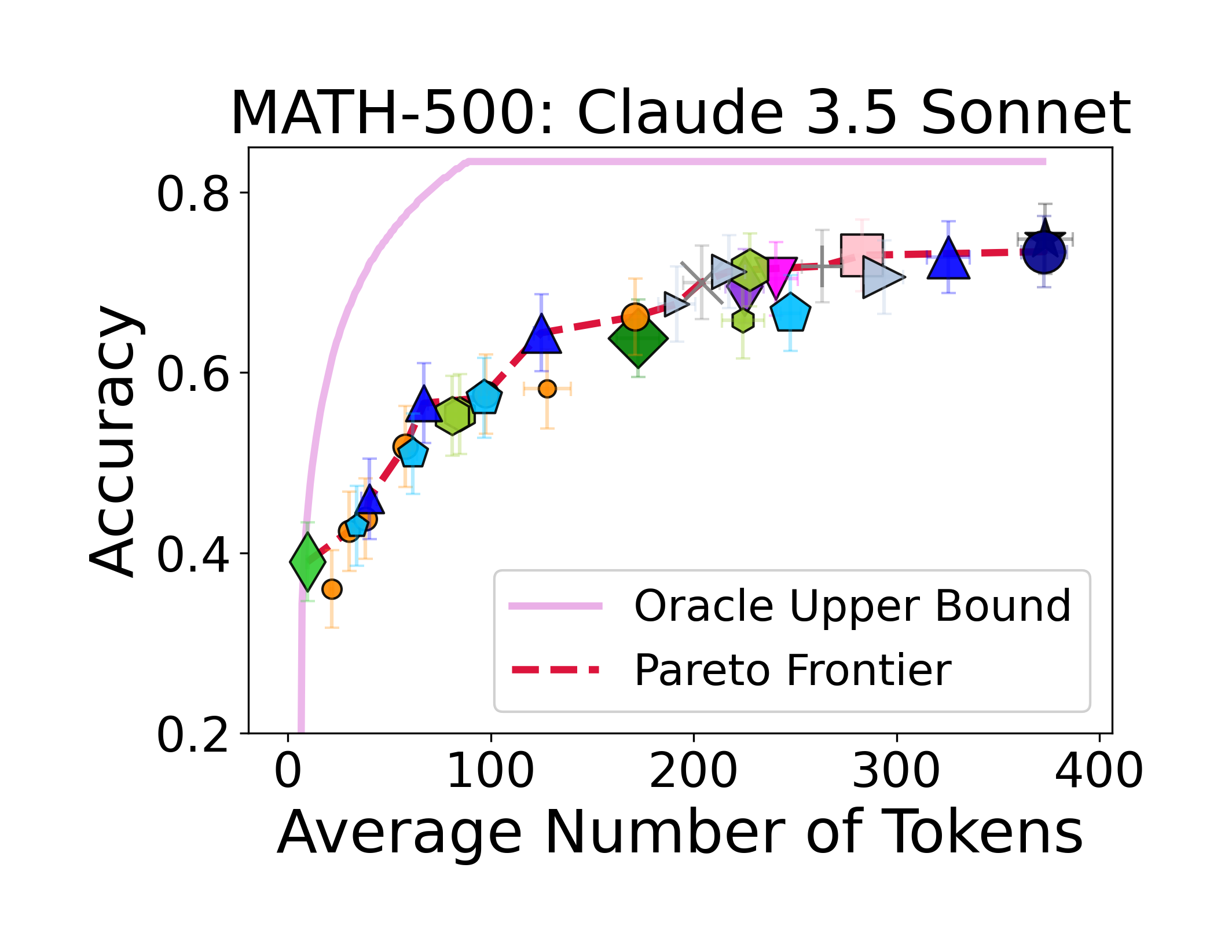}
    \includegraphics[width=0.48\linewidth]{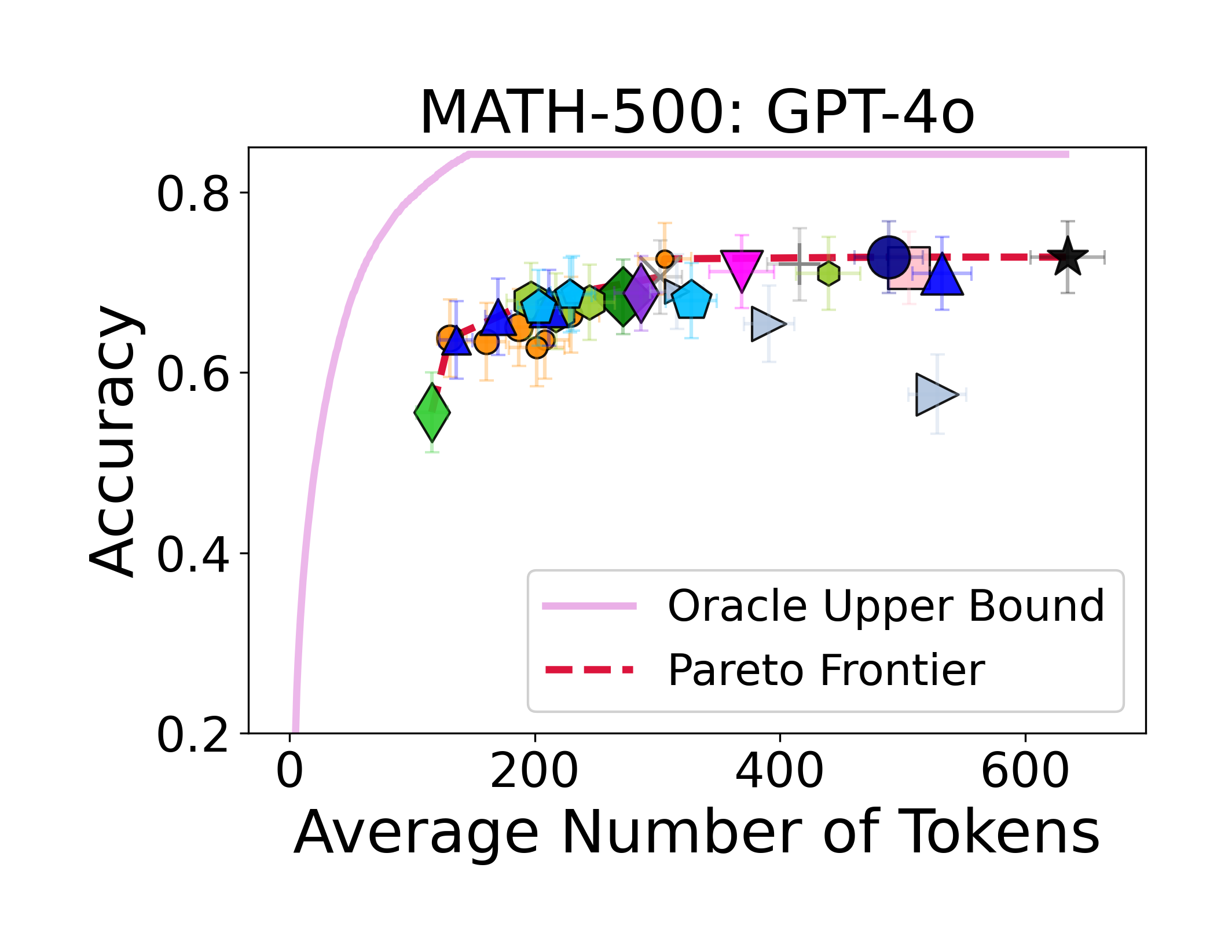}

      \includegraphics[width=0.48\linewidth]{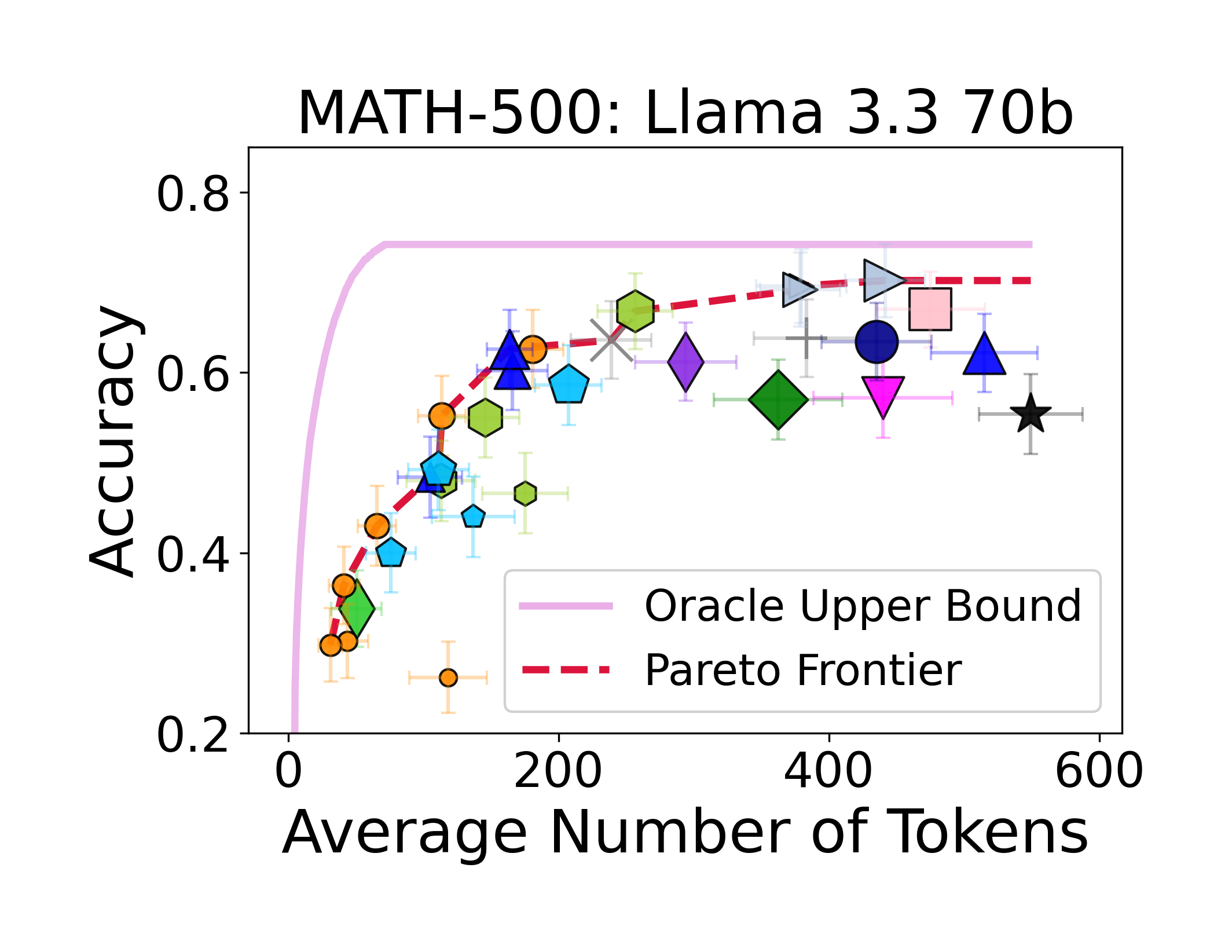}
        \includegraphics[width=0.48\linewidth]{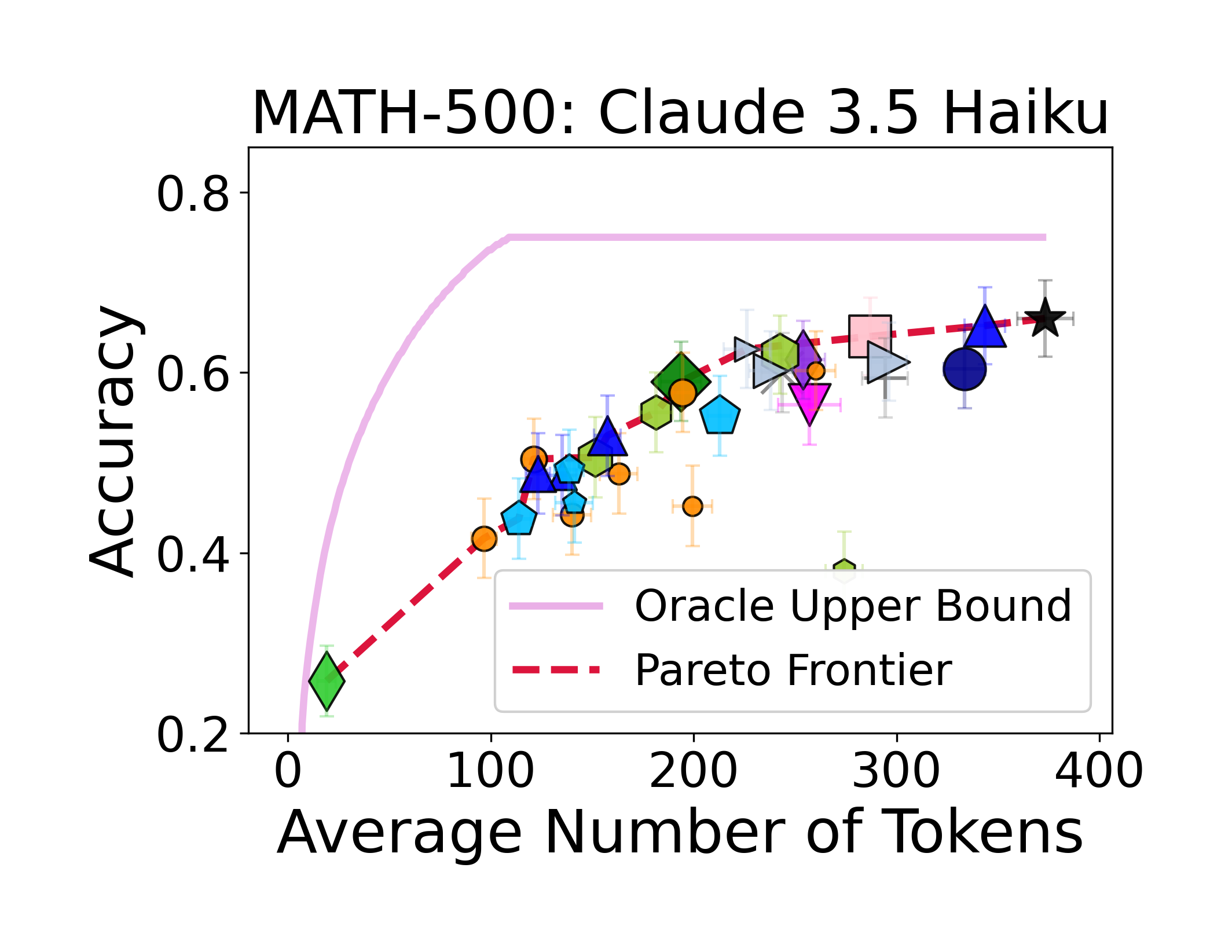}

          \includegraphics[width=0.48\linewidth]{./plot/openai/gpt-4o-mini-2024-07-18-gsm8k-main.png}
  \captionsetup{aboveskip=1pt, belowskip=2pt}
  \caption{Tradeoff Curves for  MATH-500}
  \label{fig:MATH500_ClaudePareto}
  \end{figure}

\clearpage
\section{Actual accuracy vs Predicted Accuracy from Token Complexity}
\label{sec:appendix_tradeoff}


\begin{figure}[H]
  \centering
  \includegraphics[width=0.48\linewidth]{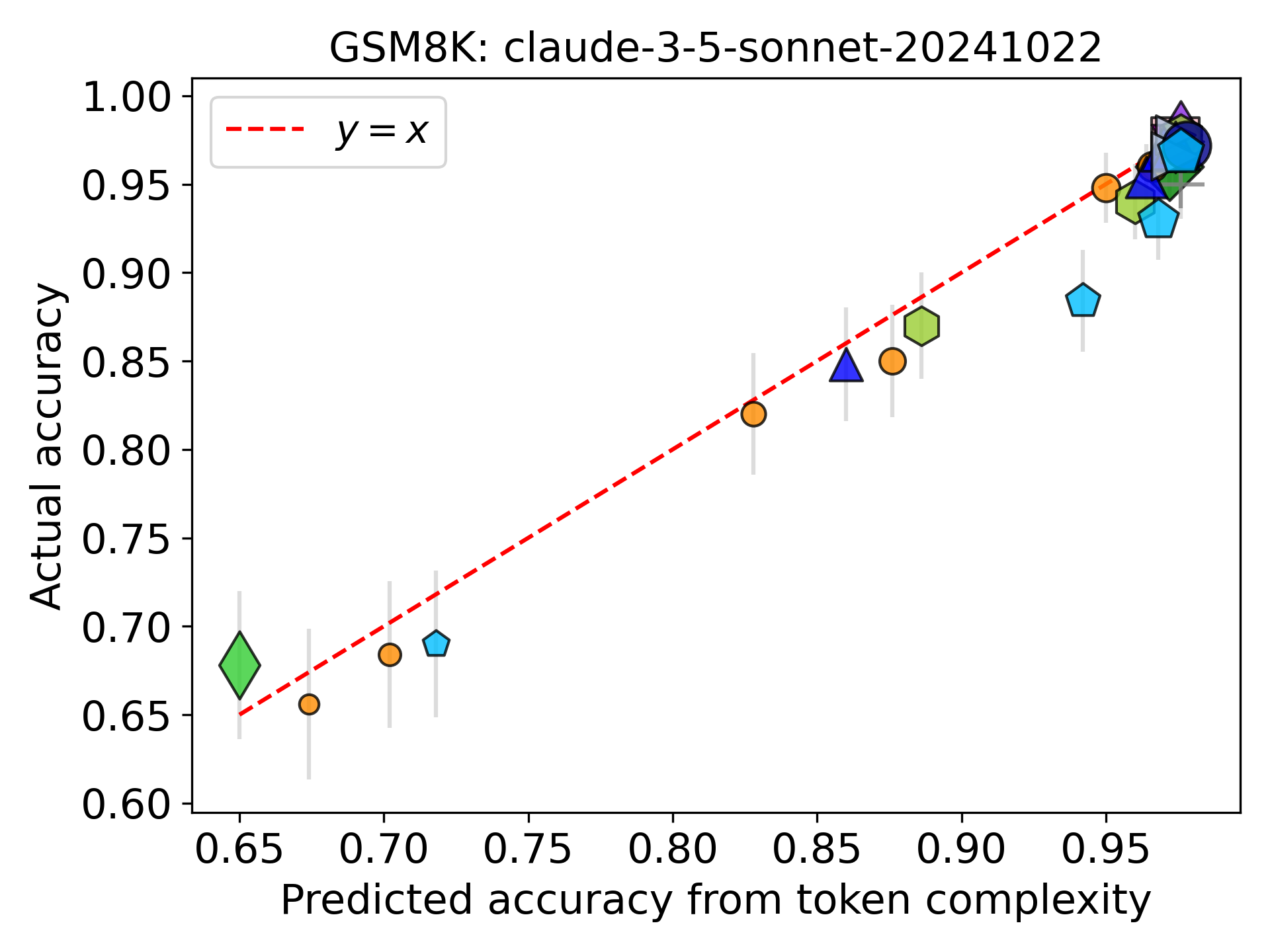}
    \includegraphics[width=0.48\linewidth]{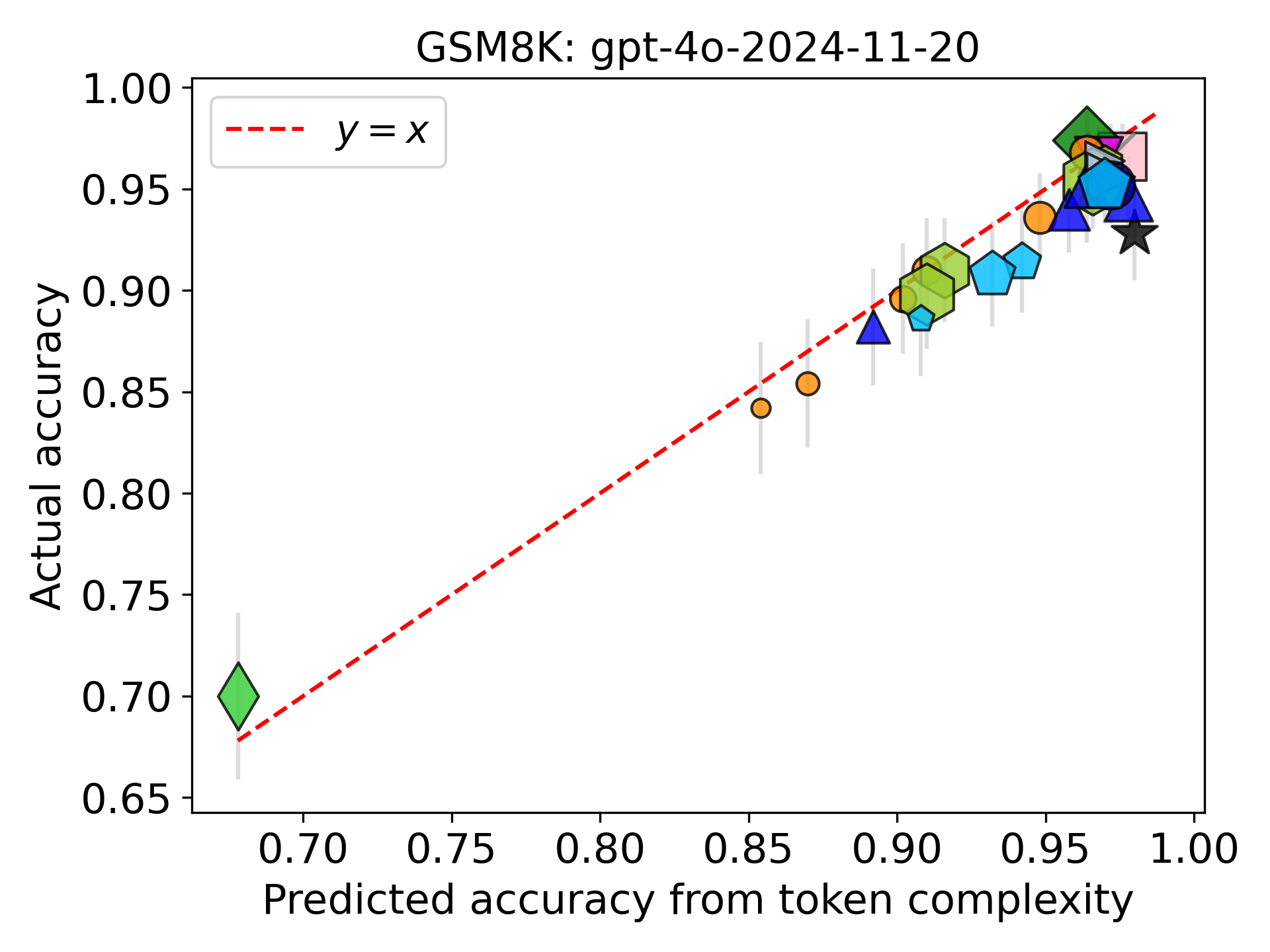}

      \includegraphics[width=0.48\linewidth]{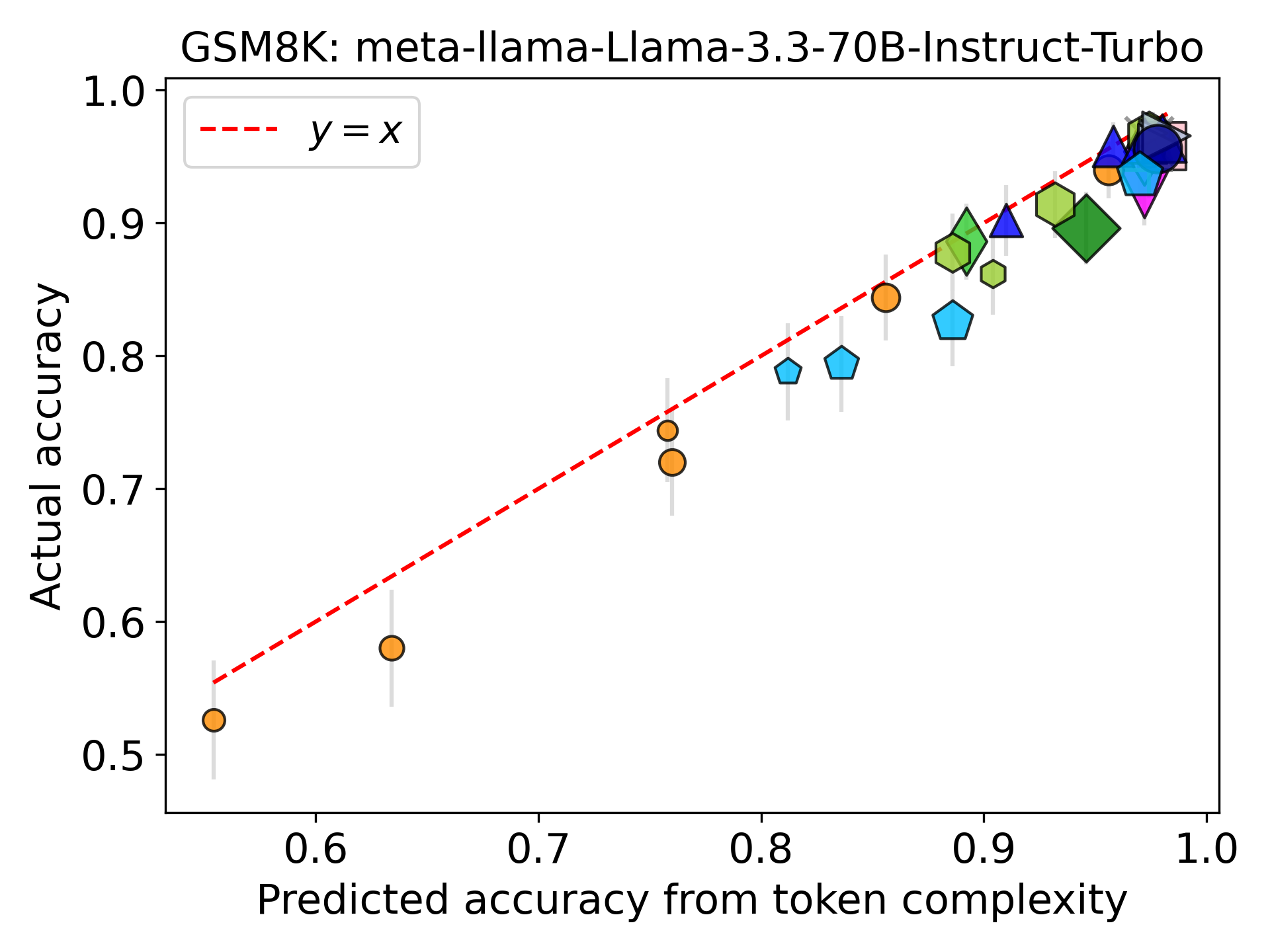}
        \includegraphics[width=0.48\linewidth]{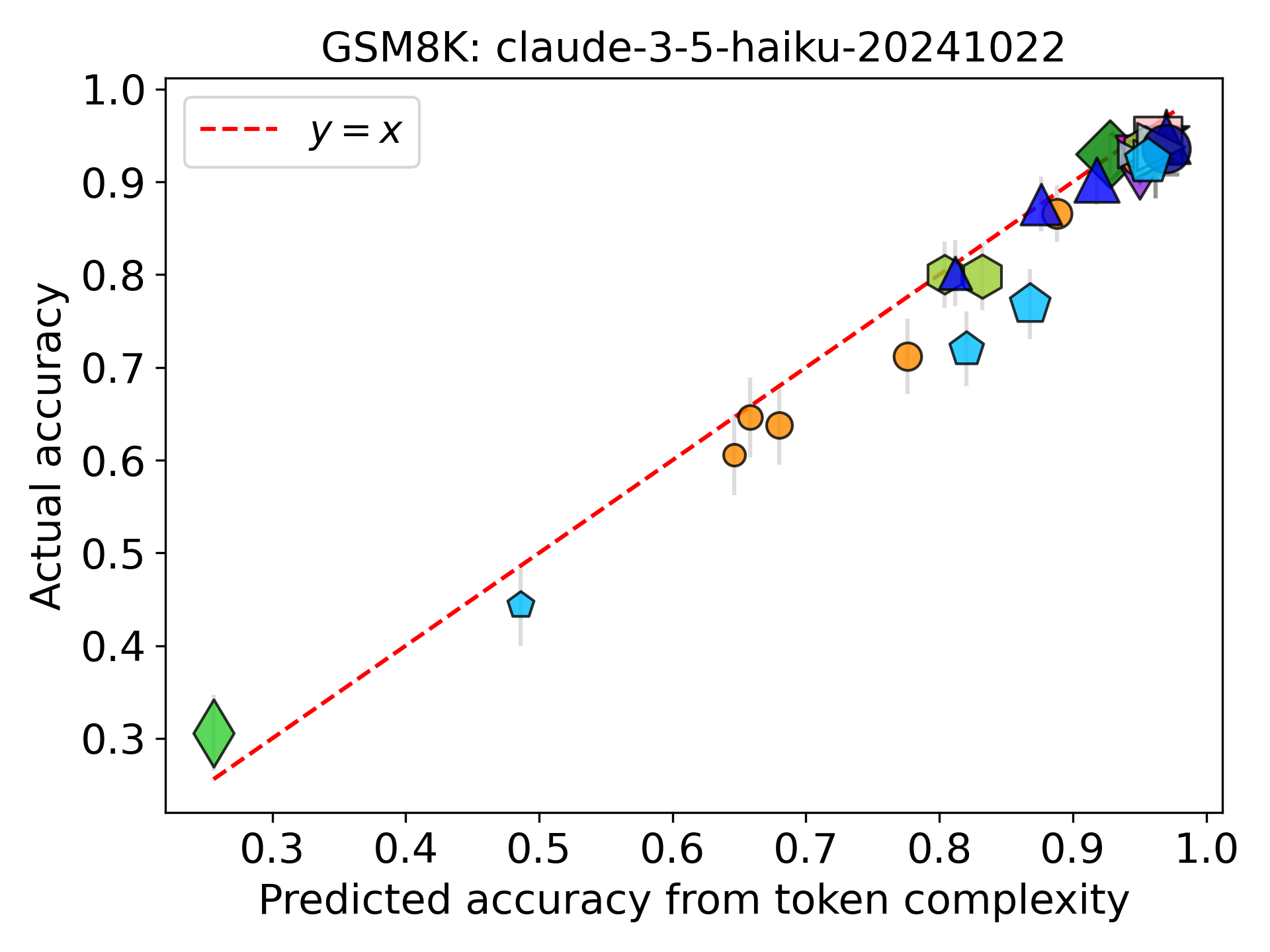}

        \includegraphics[width=0.5\linewidth]{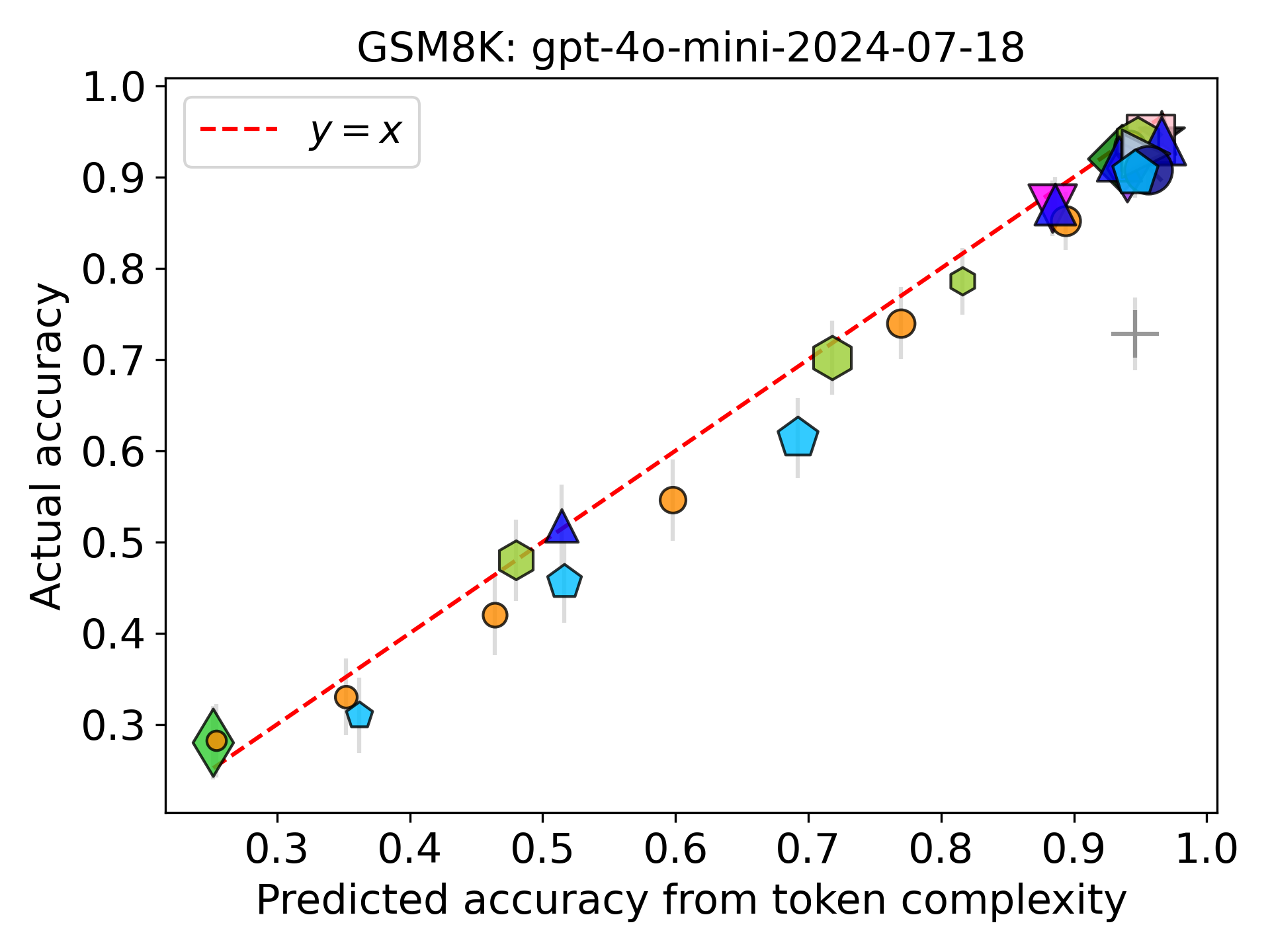}
  \captionsetup{aboveskip=1pt, belowskip=2pt}
  \caption{Actual vs Predicted Accuracy for GSM8k}
\end{figure}






\begin{figure}[H]
  \centering
  \includegraphics[width=0.48\linewidth]{./plot/Anthropic/math500_claude-3-5-sonnet-20241022_token_complexity.png}
    \includegraphics[width=0.48\linewidth]{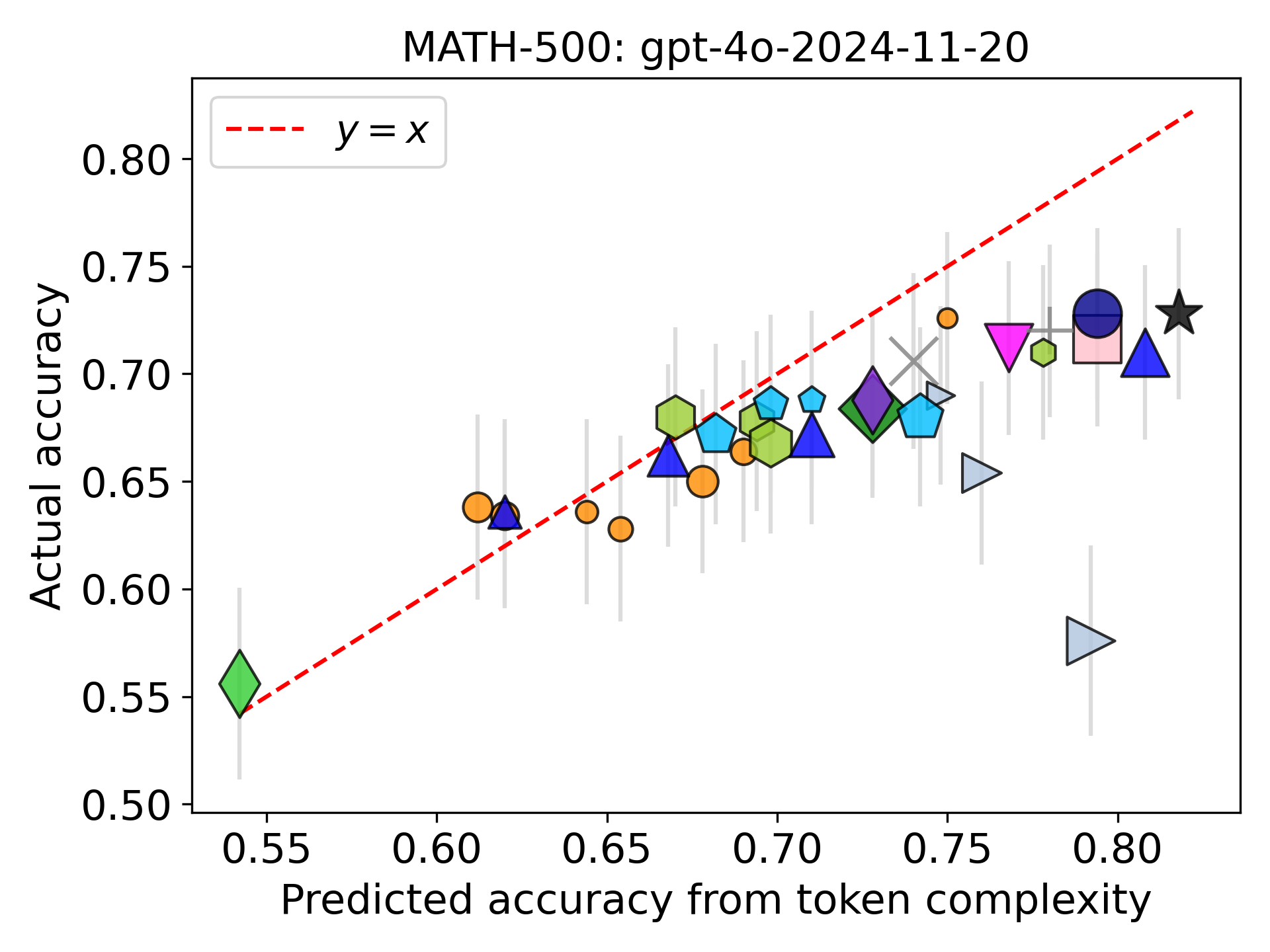}

      \includegraphics[width=0.48\linewidth]{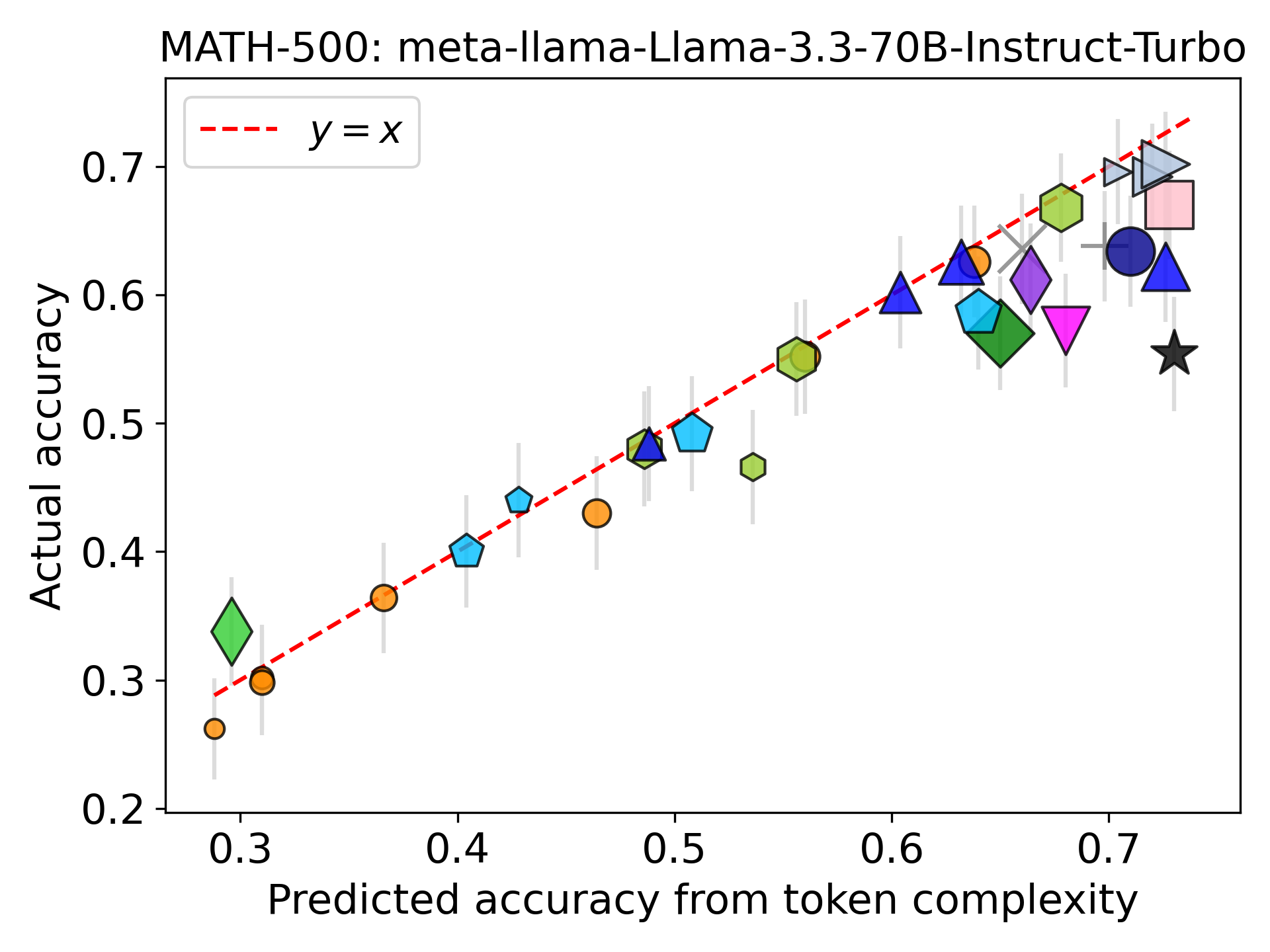}
        \includegraphics[width=0.48\linewidth]{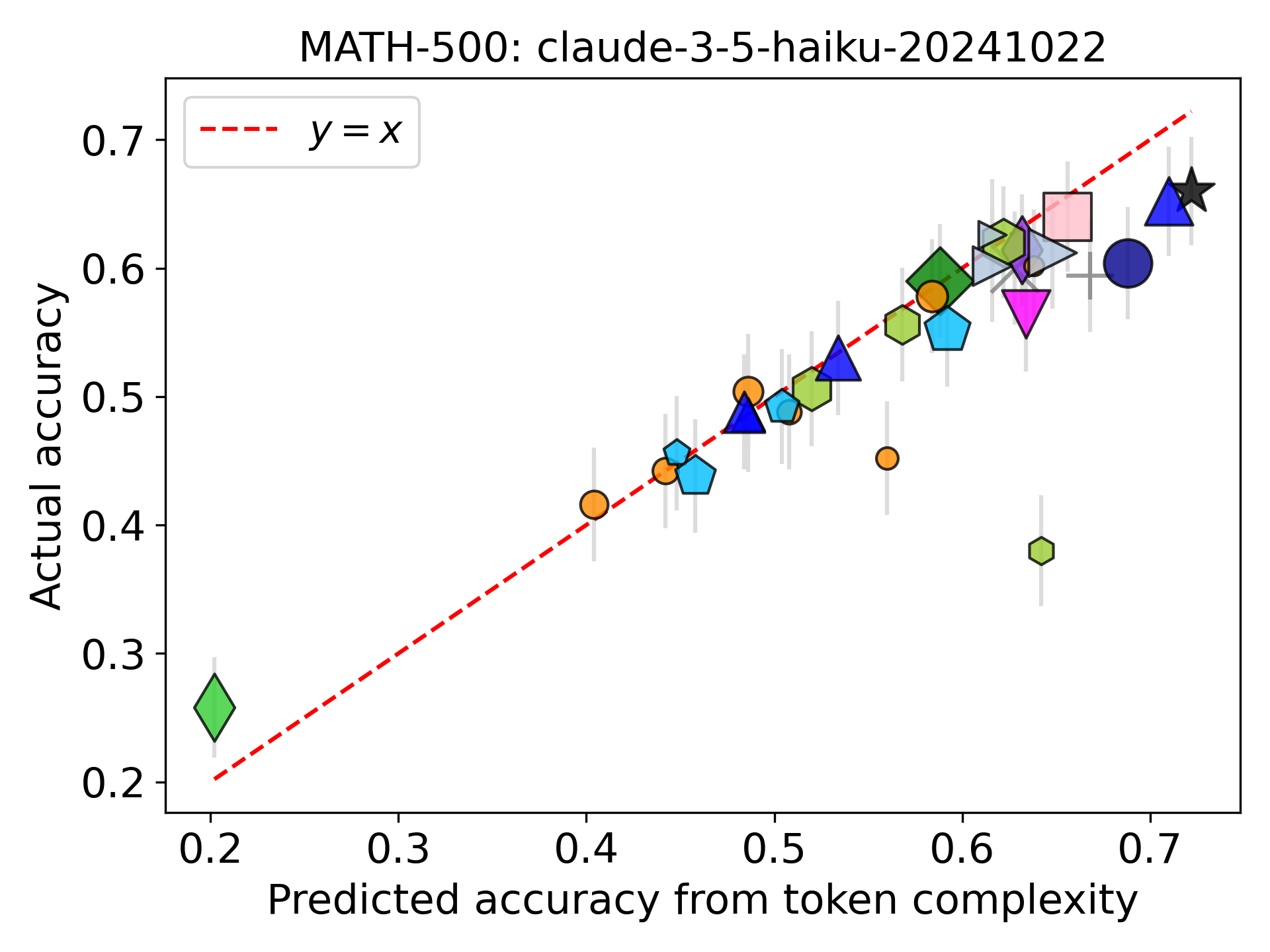}

  \includegraphics[width=0.48\linewidth]{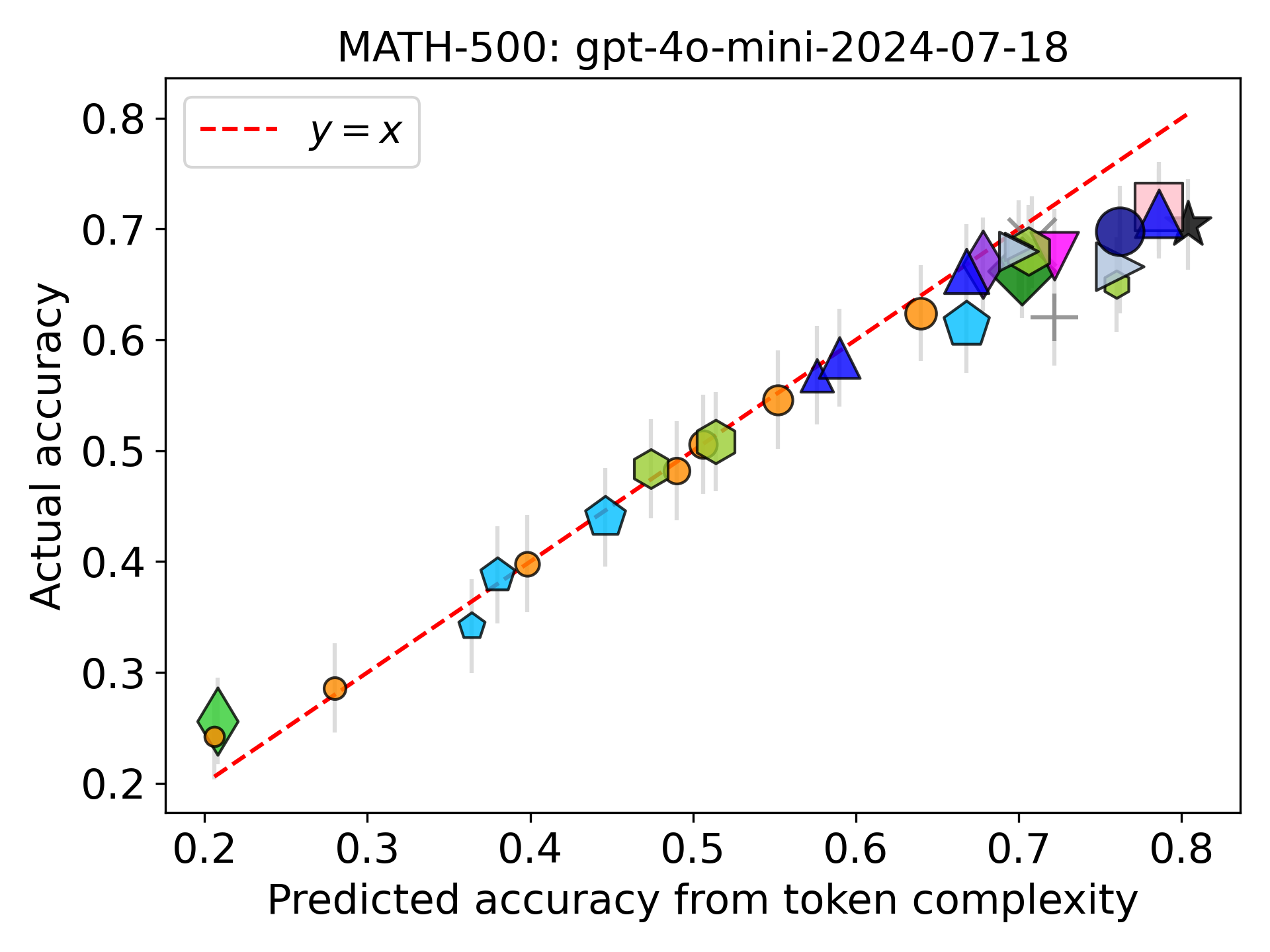}
    
  \captionsetup{aboveskip=1pt, belowskip=2pt}
  \caption{Actual vs Predicted Accuracy for MATH-500}
  \label{fig:MATH500_ClaudePrediction}
\end{figure}

\section{Routing Performance on other benchmarks}
\label{sec:appendix-routing}

\begin{figure}
  \centering
  \includegraphics[width=0.48\linewidth]{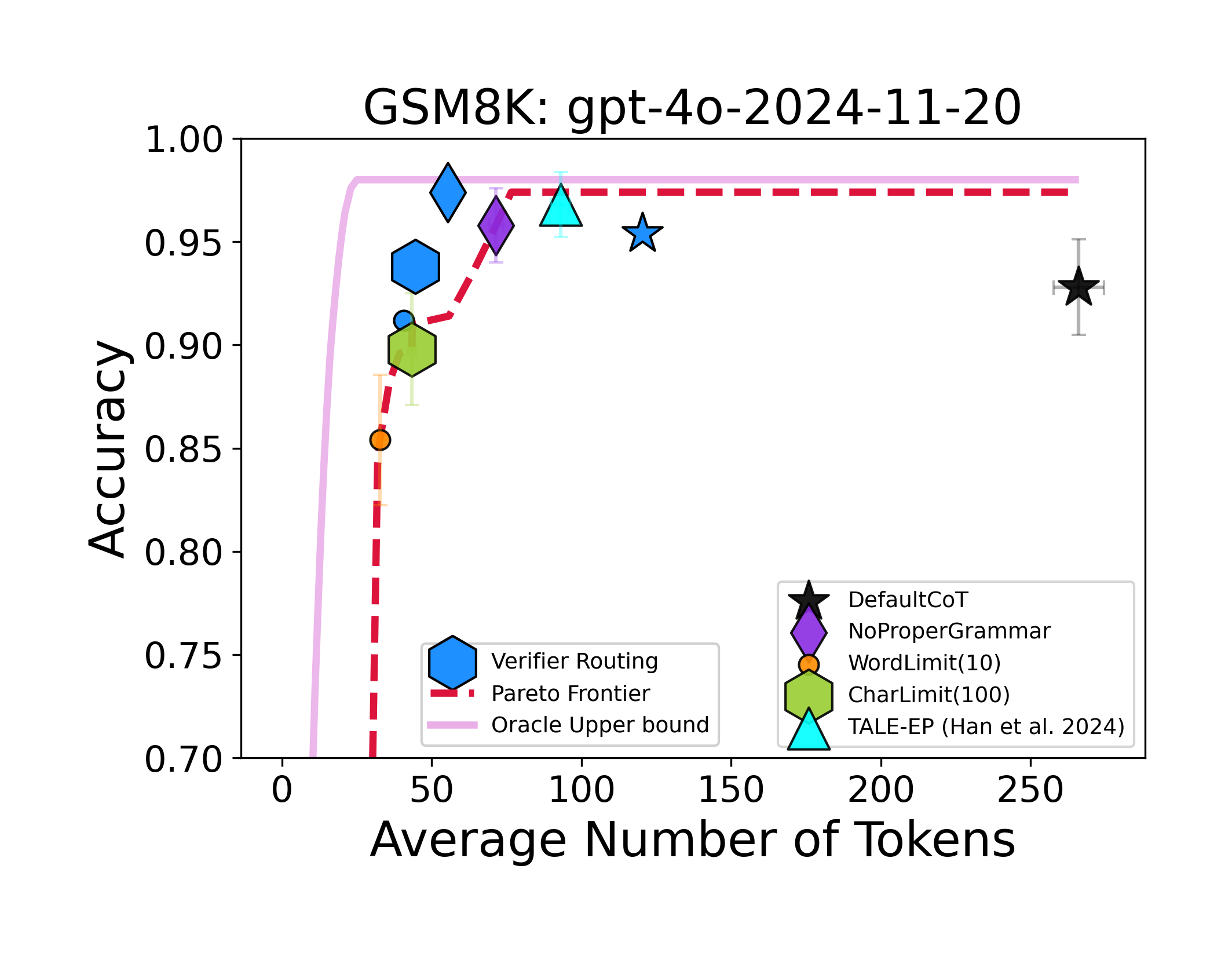}
    \includegraphics[width=0.48\linewidth]{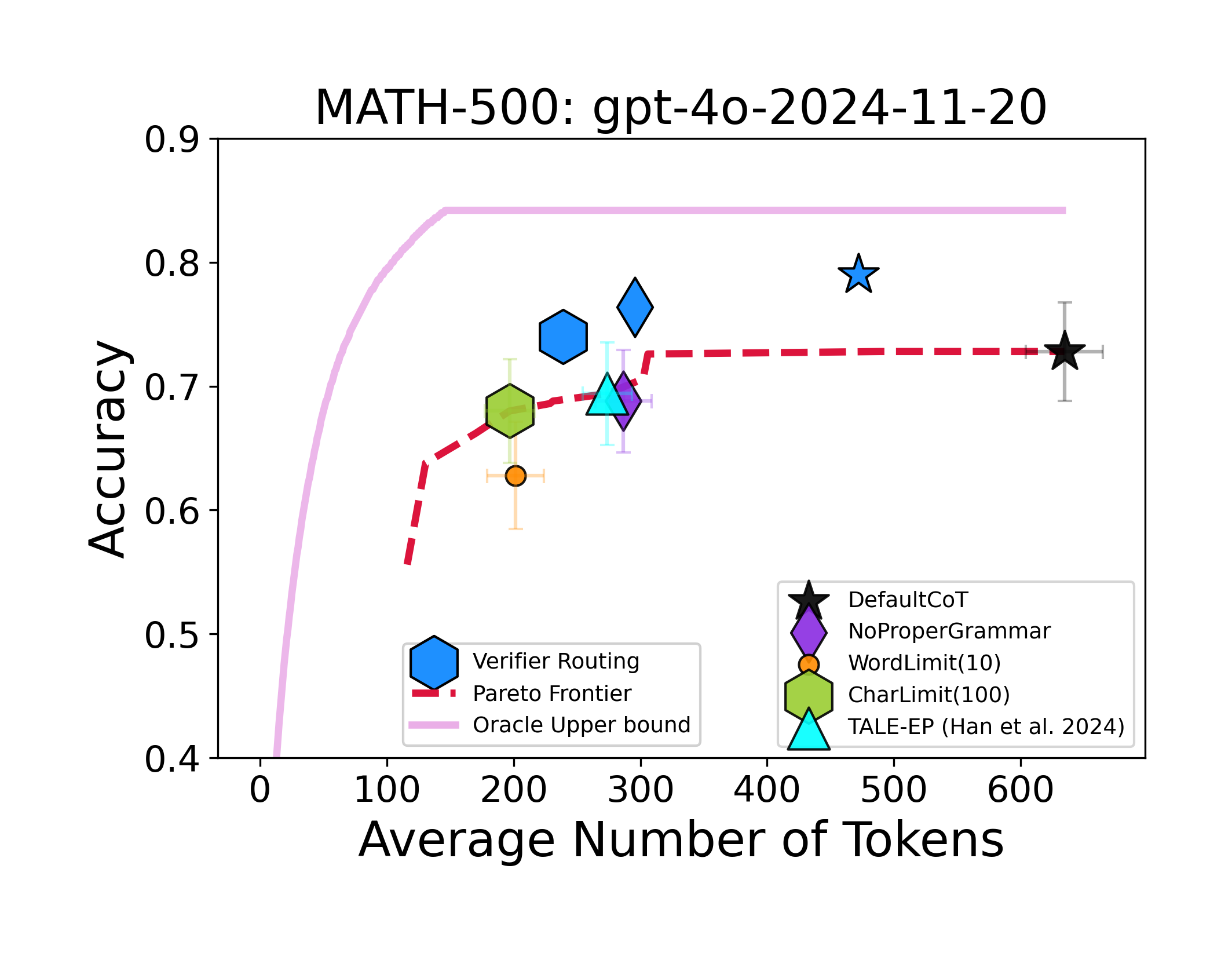}
    
  \captionsetup{aboveskip=1pt, belowskip=2pt}
  \caption{Performance of prompt routing on MATH-500 and GSM8K}
  \label{fig:prompt-routing-other}
\end{figure}

\end{document}